\documentclass{article}

\usepackage{arxiv}
\usepackage{times}
\usepackage[T1]{fontenc}
\usepackage{natbib}
\setcitestyle{authoryear,round,citesep={;},aysep={,},yysep={;}}

\newcommand{\arxivversion}{}

\usepackage{microtype}
\usepackage{graphicx}
\usepackage{subcaption}
\usepackage{booktabs} 
\usepackage{hyperref}
\usepackage{amsmath}
\usepackage{mathtools}
\usepackage{amsthm}
\usepackage{caption}
\usepackage{wrapfig}
\usepackage{enumitem}
\usepackage{array}
\usepackage{multirow}
\usepackage{colortbl}
\usepackage{xcolor}
\definecolor{revcolor}{rgb}{0.0, 0.25, 0.75}
\newif\ifshowrev
\showrevtrue
\newcommand{\rev}[1]{\ifshowrev{\color{revcolor}#1}\else#1\fi}
\DeclareCaptionFont{revcolor}{\color{revcolor}}
\newenvironment{revblock}{\ifshowrev\color{revcolor}\captionsetup{font+=revcolor}\fi}{}

\usepackage{amsmath, amssymb, amsfonts,bm, dsfont}

\usepackage{algorithm}
\usepackage{algpseudocode}

\usepackage[frozencache=true,cachedir=minted-cache]{minted} 
\setminted{style=default}

\usepackage[most]{tcolorbox}
\definecolor{myhighlight}{rgb}{1.0, 0.97, 0.8}  
\definecolor{bg}{rgb}{0.96, 0.96, 0.96}
\newenvironment{codeblock}
  {\begin{tcolorbox}[colback=bg, colframe=white, boxrule=0pt, left=0pt, right=0pt, top=2pt, bottom=2pt, enhanced]}
  {\end{tcolorbox}}
  
\theoremstyle{plain}
\newtheorem{theorem}{Theorem}[section]

\newtheorem{lemma}[theorem]{Lemma}

\theoremstyle{definition}
\newtheorem{definition}[theorem]{Definition}
\newtheorem{assumption}[theorem]{Assumption}
\theoremstyle{remark}
\newtheorem{remark}[theorem]{Remark}

\DeclareMathAlphabet{\mathsfit}{\encodingdefault}{\sfdefault}{m}{sl}
\SetMathAlphabet{\mathsfit}{bold}{\encodingdefault}{\sfdefault}{bx}{n}

\DeclareMathOperator*{\bE}{\mathbb{E}}

\def\bP{{\mathbb{P}}}

\def\bR{{\mathbb{R}}}
\def\bS{{\mathbb{S}}}

\def\bV{{\mathbb{V}}}

\def\bX{{\mathbb{X}}}
\def\bY{{\mathbb{Y}}}

\def\indicator{{\mathds{1}}}

\def\cA{{\mathcal{A}}}

\def\cG{{\mathcal{G}}}

\def\cJ{{\mathcal{J}}}
\def\cK{{\mathcal{K}}}
\def\cL{{\mathcal{L}}}

\def\cT{{\mathcal{T}}}

\newcommand{\pdata}{p_{\rm{data}}}

\makeatletter
\def\Ldiffuse{\@ifnextchar_{\Ldiffuse@}{\mathcal{L}^{\text{diffuse}}}}

\def\Ldenoise{\@ifnextchar_{\Ldenoise@}{\mathcal{L}^{\text{denoise}}}}
\def\Ldenoise@_#1{\mathcal{L}_{#1}^{\text{denoise}}}

\def\Lmask{\@ifnextchar_{\Lmask@}{\mathcal{L}^{\text{mask}}}}
\def\Lmask@_#1{\mathcal{L}_{#1}^{\text{mask}}}

\def\Lunmask{\@ifnextchar_{\Lunmask@}{\mathcal{L}^{\text{unmask}}}}
\def\Lunmask@_#1{\mathcal{L}_{#1}^{\text{unmask}}}

\def\Ldelete{\@ifnextchar_{\Ldelete@}{\mathcal{L}^{\text{delete}}}}
\def\Ldelete@_#1{\mathcal{L}_{#1}^{\text{delete}}}

\def\Linsert{\@ifnextchar_{\Linsert@}{\mathcal{L}^{\text{insert}}}}
\def\Linsert@_#1{\mathcal{L}_{#1}^{\text{insert}}}

\def\len{{\text{len}}}

\def\delete{{\text{delete}}}
\def\simple{{\text{simple}}}
\def\data{{\text{data}}}
\def\ins{{\text{ins}}}

\newcommand{\abbfid}{FID$\downarrow$}
\newcommand{\abbmiou}{mIoU$\uparrow$}
\newcommand{\abbalignment}{Ali$\rightarrow$}
\newcommand{\abboverlap}{Ove$\rightarrow$}
\newcommand{\taskugen}{Un-Gen}
\newcommand{\tasktype}{Gen-Type}
\newcommand{\tasktypesize}{Gen-TypeSize}
\newcommand{\taskcompletion}{Completion}
\newcommand{\taskrefinement}{Refinement}

\showrevfalse 

\usepackage{hyperref}
\usepackage{url}
\hypersetup{colorlinks=true, linkcolor=red!60!black, citecolor=blue!60!black, urlcolor=magenta!70!black}

\title{Variable Length Diffuse Everything: Arbitrary Length Generative Modeling on Arbitrary State Spaces}

\renewcommand{\shorttitle}{Variable Length Diffuse Everything}

\author{
  Kevin Rojas \\
  Georgia Institute of Technology \\
  \texttt{kevin.rojas@gatech.edu} \\
  \And
  Felix X.-F.~Ye \\
  University at Albany, SUNY \\
  \texttt{xye2@albany.edu} \\
  \And
  Molei Tao \\
  Georgia Institute of Technology \\
  \texttt{mtao@gatech.edu} \\
}

\date{}

\begin{document}

\maketitle

\begin{abstract}
Diffusion and flow-based generative models have achieved remarkable results across diverse applications, but most assume data of fixed-dimension/length. Meanwhile, real-world problems such as layout generation often involve variable-length, multimodal data, such as continuous bounding box coordinates paired with discrete types. Existing methods typically address variable length by padding data to a fixed size or by introducing specialized objectives to model lengths explicitly, adding complexity and limiting generality.
We propose Variable Length Diffuse Everything (VL-DE), a simple and general framework for variable-length multimodal generation on arbitrary state spaces. Instead of padding, VL-DE handles variable length through an insertion-based mechanism grounded in a principled theoretical framework, yielding a tractable training objective. In addition, by utilizing the fact that VL-DE is based on a generator with an additive form, we develop a higher-order sampler by appropriately handling the time dependence. Extensive experiments on layout generation demonstrate the effectiveness of our approach.
\end{abstract}

\section{Introduction}
Generative models have led to significant advances across a wide range of scientific and practical applications. Diffusion and flow-based models have been established across a broad range of settings, including Euclidean spaces \citep{song2020score, ho2020denoising, albergo2023stochastic, lipman2022flow, liu2022flow}, 
curved manifolds \citep{de2022riemannian, huang2022riemannian, chenflow, zhutrivialized}, discrete spaces \citep{campbell2022continuous, lou2023discrete, ou2024your, sahoo2024simple, shi2024simplified}, and more recently, multimodal data \citep[e.g.,][]{holderrieth2024generator, campbell2024generative, rojas2025diffuse, benton2024denoising, ren2025unified}.
Variable-length generation has likewise received substantial attention, but almost exclusively in single-modality settings \citep{campbell2023trans, kim2025any, havasi2025edit}.
Multimodal generation, on the other hand, has been studied predominantly at fixed length.
Only a handful of works sit at the intersection of these two lines \citep{billera2025branching, nguyen2025oneflow}, and even these impose strong structural constraints; such as tying all modalities to a single length.

Variable-length multimodal data arises naturally in many domains. One such example is layout generation, where a layout is a variable-length sequence of discrete box types paired with continuous box coordinates, and both the number of boxes and their arrangement must be generated jointly.
\begin{figure}[t]
\vspace{-\baselineskip}
\centering
\includegraphics[width=\linewidth]{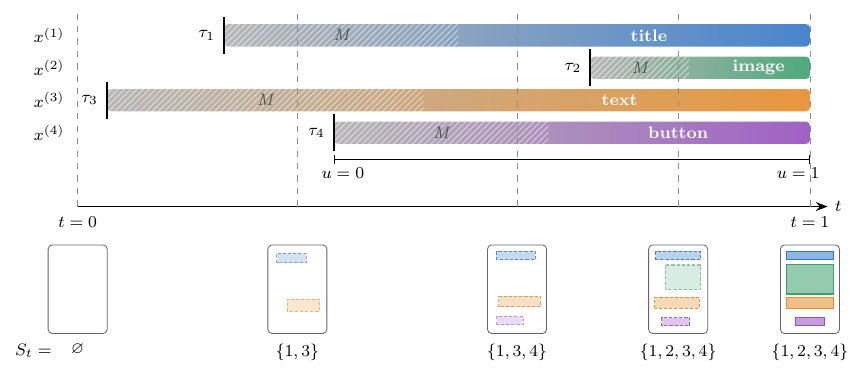}
\caption{
Overview of \textbf{VL-DE} on a layout example. Variable-length data is generated by a joint process $(X_t, S_t)$, where $S_t$ is an \emph{auxiliary process} that tracks the active coordinates and $X_t$ carries their values. Each coordinate $i$ receives an insertion time $\tau_i$, an insertion distribution and is denoised on its own clock $u$, so that all elements are clean simultaneously at $t=1$.
}
\vspace{-\baselineskip}
\vspace{-\baselineskip}
\label{fig:paper-overview}
\end{figure}
A key challenge in such settings is that length is rarely known a priori: much like a language model does not know how many tokens it will produce before generation begins, the appropriate length only becomes apparent as the content unfolds. This makes length an implicit, content-dependent quantity, and suggests that decoupled approaches — in which one first samples a length and then samples data conditioned on it — are unlikely to capture the joint structure faithfully.

A standard approach to variable-length generation relies on auto-regressive models, which generate sequences through a series of insertions.
While effective in many settings, auto-regressive generation imposes a left-to-right factorization that is not always natural for the underlying data distribution. This has motivated work to extend diffusion and flow-based models to support variable-length data.
\emph{Unimodal} methods such as \emph{Transdimensional Diffusion} \citep{campbell2023trans}, \emph{FlexMDM} \citep{kim2025any}, and \emph{Edit Flows} \citep{havasi2025edit} extend diffusion or flow-based models with insertion or editing mechanisms, but do not address the multimodal case.
\emph{Interleaving-based multimodal} methods such as \emph{Transfusion} \citep{zhoutransfusion} and \emph{OneFlow} \citep{nguyen2025oneflow} handle the specific pairing of discrete tokens with continuous values by interleaving them in a single sequence; this construction is tied to that particular pairing and does not extend to general orderings in general state spaces.
\emph{General-state-space multimodal} frameworks such as \emph{Branching Flows} \citep{billera2025branching} allow arbitrary state spaces, but their duplication dynamics may not suit all data, and they impose structural restrictions like having all modalities share a length. 

In this work, we introduce \emph{Variable Length Diffuse Everything} (VL-DE), a framework whose construction rests on two frameworks: \emph{Generator Matching} \citep{holderrieth2024generator} and \emph{Diffuse Everything} \citep{rojas2025diffuse}. Generator Matching offers a general recipe to learn generators of continuous-time Markov processes on arbitrary state spaces, while Diffuse Everything shows how to model multimodal data by combining per-modality generators under decoupled time variables. Neither, however, accommodates variable-length data: Generator Matching applied directly leads to intractable training objectives (as in \citet{campbell2023trans}), and Diffuse Everything inherits these issues.

VL-DE addresses both limitations. We extend Generator Matching by augmenting the state space with an auxiliary variable that records the latent operations producing a state. Marginalizing this variable yields a generator on the original variable-length data space, while training on the joint augmented process keeps the objective tractable. As a concrete instantiation, we specify the training process directly as a \emph{variable-length interpolant}: each coordinate of a data point is assigned an insertion time, then as in Diffuse Everything, each coordinate is denoised on its own clock, so that all coordinates are clean simultaneously at the end (see Figure~\ref{fig:paper-overview}). The resulting generator combines insertion and denoising on arbitrary state spaces, and recovers many  variable length approaches such as \citep{kim2025any, havasi2025edit, nguyen2025oneflow}.

Beyond the design of the generative process, efficient \emph{sampling} is essential for practical deployment. To this end, we leverage the fact that our approach is based on the generator of a Markov process that is naturally the sum of multiple terms, and develop higher-order samplers. Notably, traditional operator splitting techniques~\citep{mclachlan2002splitting} do not directly apply because each term is time dependent, and we introduce a novel non-autonomous operator splitting scheme for general Markov processes. These samplers improve sample quality per step at small step counts.

We evaluate VL-DE on variable-length multimodal generation on both unconditional and conditional layout generation. Experimental evidence demonstrates that VL-DE can perform on par with even oracle-length models without sacrificing sample quality and beats other variable length methods.

Our contributions are summarized as follows:

\begin{itemize}[leftmargin=*, noitemsep, topsep=0pt, parsep=0pt]
    \item For generatively modeling multimodal data with variable dimensions, we develop a principled construction based on augmenting state space with auxiliary variables, and show that the resulting dynamics can be marginalized to obtain generators defined on the original state space.

    \item We instantiate this construction as a variable-length interpolant with insertion and per-coordinate clocks, which is defined on arbitrary state spaces, is trained without marginalizing over latent insertion histories, and can handle multiple lengths.

    \item We develop an efficient sampling method based on a non-autonomous operator splitting for general Markov processes, improving accuracy per step in the large step size regime.

\end{itemize}

\section{Background}
\paragraph{Generator Matching} Many generative models construct samples by interpolating between a simple reference distribution $p_{\simple}$ and the data distribution $p_{\data}$. This includes diffusion, flow-based, and related models across Euclidean, discrete, and multimodal state spaces. A unifying perspective is to view these models as continuous-time Markov processes. Concretely, given a state space $\bX$ and data distribution $p_{\data}$, one specifies a \emph{conditional probability path} $p_{t|1}(dx \mid x_1)$ for $t\in[0,1]$ satisfying
\[
p_{1|1}(dx \mid x_1) = \delta_{x_1}(x), 
\qquad 
p_{0|1}(dx \mid x_1) = p_{\simple}.
\]
Averaging over $x_1 \sim p_{\data}$ yields the marginal path
$
p_t(dx) = \mathbb{E}_{x_1 \sim p_{\data}}\!\left[p_{t|1}(dx \mid x_1)\right],
$
which interpolates between $p_0 = p_{\simple}$ and $p_1 = p_{\data}$. If $(p_t)_{t\in[0,1]}$ is generated by a (possibly time-inhomogeneous) Markov process with generator $\cL_t$ (refer to Appendix \ref{app:background-generators} for a formal introduction), then its evolution satisfies the weak Fokker-Planck equation
\begin{align}\label{eq:weak-fokker-planck}
\partial_t \langle p_t, f \rangle = \langle p_t, \cL_t f \rangle,
\end{align}
for suitable test functions $f$. Thus, learning the generator $\cL_t$ is sufficient to specify the generative process. Generator Matching \citep{holderrieth2024generator} provides a general framework for learning such generators. Suppose that for each data point $x_1$, there exists a conditional generator $\cL_{t,x_1}$ generating the path $p_{t|1}(\cdot \mid x_1)$. Then the generator of the marginal path admits the factorization
\begin{align}\label{eq:generator-factorization}
\cL_t f(x) = \mathbb{E}_{x_1 \sim p_{1|t}(\cdot \mid x)}\!\left[\cL_{t,x_1} f(x)\right],
\end{align}
where $p_{1|t}(\cdot \mid x)$ denotes the posterior over data points given the intermediate state $x$ at time $t$. Furthermore, for a broad class of Markov processes, the generator admits a linear parameterization
\begin{align}\label{eq:generator-linear}
\cL_t f(x) = \langle K f(x), F_t(x) \rangle,
\end{align}
where $K$ is a fixed linear operator and $F_t(x)$ characterizes the dynamics. Generator Matching aims at learning a neural approximation $F_t^\theta$ by minimizing
\begin{align}\label{eq:generator-matching-loss}
L_{\mathrm{gm}}(\theta)
= \mathbb{E}_{t} \, \mathbb{E}_{x_t \sim p_t}
\big[ D\!\left(F_t(x_t), F_t^\theta(x_t)\right) \big],
\end{align}
where $D$ is a Bregman divergence. Since this objective is generally intractable, one instead minimizes the \emph{conditional generator matching} objective
\begin{align}\label{eq:conditional-generator-matching}
L_{\mathrm{cgm}}(\theta)
= \mathbb{E}_{t} \, \mathbb{E}_{x_1 \sim p_{\data}} \, \mathbb{E}_{x_t \sim p_{t|1}(\cdot \mid x_1)}
\big[ D\!\left(F_{t,x_1}(x_t), F_t^\theta(x_t)\right) \big],
\end{align}
where $F_{t,x_1}$ parameterizes $\cL_{t,x_1}$. A key result in \citet{holderrieth2024generator} shows that minimizing this objective recovers the generator of the marginal path.



\paragraph{Diffuse Everything} \citet{rojas2025diffuse} models multimodal data on arbitrary state spaces. It was originally formulated in the Denoising Markov Models framework \citep{benton2024denoising, ren2025unified}, where the generative process is reversing a forward noising process; however, its core construction applies more broadly to any number of Markov processes specified by their generators, including those that do not arise from a forward process. We adopt this more general view here.

Given two modalities $x \in \bX$ and $y \in \bY$, each is equipped with its own generator, $\cL^x$ and $\cL^y$, defined on its respective space. These per-modality generators are then combined into a joint generator using \emph{decoupled time variables} $(t_1, t_2) \in [0,1]^2$, one per modality:
\[
    \cL_{t_1, t_2} = \cL_{t_1}^x + \cL_{t_2}^y.
\]
Under this construction each modality evolves independently under its own clock. The decoupled times allow a single model to support multiple generation modes: setting $t_1 = t_2$ corresponds to joint generation of $(x, y)$, while holding one clock fixed (e.g., $t_1 = 1$) and running the other yields conditional generation (e.g., $y \mid x$). The framework extends to any number of modalities and state spaces. A key restriction, however, is that Diffuse Everything assumes each modality lives in a \emph{fixed-length} state space; lifting this restriction is the central goal of VL-DE. Appendix~\ref{app:de-gm} discusses this view of Diffuse Everything and the effect of decoupled times on training.


\section{Methodology}
In principle, generator matching can model variable length data by defining a Markov process over $\bX = \bigcup_{d=1}^{D} \tilde\bV^d$, where $\tilde\bV$ is an arbitrary space that may itself be multimodal (e.g.,\ $\tilde\bV = \bR^4\times[N]$). 
If this process changes dimensions (e.g., through insertions), it can model variable-length data --- as in Trans-dimensional Diffusion \citep{campbell2023trans}.
However, the difficulty lies in the conditional probability path: many latent operations (e.g., different insertion paths) can produce the same state, so evaluating the loss function requires summing over all such possibilities. This is generally intractable, and \citet{campbell2023trans} relies on a practical simplification to make training feasible. 

To sidestep this issue, \emph{Edit Flows} \citep{havasi2025edit} and \emph{FlexMDM} \citep{kim2025any} concurrently introduced an auxiliary process that explicitly tracks the latent operations producing each state, removing the ambiguity and making training tractable. However, both constructions restrict the data and the auxiliary process to discrete spaces,
which limits their applicability to multimodal data.

We address this in two steps. Section~\ref{sec:generator-matching-with-auxiliary-process} generalizes the auxiliary-process idea to arbitrary state spaces, yielding \emph{Generator Matching with an Auxiliary Process}; this already covers multimodal data with a common length. Section~\ref{sec:practical-process} then instantiates it as a variable-length interpolant in which every coordinate is denoised on its own clock; since a coordinate may itself be a variable-length object, this also covers modalities with \emph{independently varying lengths}, a regime not handled by prior work.

\subsection{Generator Matching with an Auxiliary Process} \label{sec:generator-matching-with-auxiliary-process}
 \noindent
\textbf{Notation.}
We use $\cL$ to denote generators acting on augmented state spaces (e.g., including auxiliary variables), and $\cK$ to denote generators acting on the marginal data space.
\paragraph{Setup:} We start by considering a distribution $p_\text{data}$ over a space $\bX$ that we want to generate new samples from. Unlike standard approaches that define a generator on $\bX$, we work on a joint state space $\bX\times \bS$, where the auxiliary space $\bS$ records the latent operations that produced a corresponding state in $\bX$ --- for example, in Edit Flows the auxiliary process keeps track of the sequence of edit operations used. We require this generator to satisfy the same principle as in Equation \ref{eq:generator-factorization}:
\begin{align} \label{eq:joint-generator-factorization}
    \cL_t f(x,s) &= \bE_{x_1, s_1}\left[ \cL_{t,x_1, s_1} f(x,s) | (X_t,S_t) = (x,s)\right]
\end{align}

Furthermore, we assume that both $\cL_t$ and $\cL_{t,x_1,s_1}$ can be parametrized linearly as:
\begin{align}\label{eq:joint-generator-linear}
    \cL_t f(x,s) = \langle K f(x,s), F_t(x,s) \rangle, \quad \cL_{t,x_1,s_1} f(x,s) = \langle K f(x,s), F_{t,x_1,s_1} (x,s) \rangle
\end{align}
are the same conditions as in Generator Matching~\citep{holderrieth2024generator}, reviewed in \eqref{eq:generator-factorization} and \eqref{eq:generator-linear}, except now on the extended state space. These conditions allow defining a generative model over $\bX \times \bS$. Naively doing so would force us to model $S$, leading to an unnecessary computational overhead. We propose a better alternative, while keeping the benefits of the auxiliary process.

\paragraph{The marginalized Process:} The key insight is that the role of $S$ is to make training tractable by removing the ambiguity of many auxiliary sequences producing the same $x$.
We therefore construct a generator on $\bX$ that \emph{marginalizes} $S$ while still letting us exploit it during training, as formalized by the following theorem:

\begin{theorem}\label{thm:joint-generators}
    \rev{Under the regularity assumptions of Appendix~\ref{app:aux-gen-proofs},} let $\cL_t f(x,s)$ satisfy equations \eqref{eq:joint-generator-factorization} and \eqref{eq:joint-generator-linear} as well as the weak Fokker-Planck equation \eqref{eq:weak-fokker-planck}, and let $\varphi_f(x,s) := f(x)$
    denote the lift of $f$ to $\bX\times\bS$. Define the generator $\cK_t$ on $\bX$ by
    \begin{align}
        \cK_t f(x) = \bE_{s_t} [ \cL_t \varphi_f(x,s_t) \mid X_t = x].
    \end{align}
    Then $\cK_t$ generates the marginal probability path $p_t(x)$.
    Furthermore, the generator admits a factorization given by:
    \begin{align}
         \cK_t f(x) = \bE_{x_1, s_1, s_t} \left[\cK_{t,x_1, s_1, s_t} f(x) | X_t = x \right]
    \end{align}
    where the conditional generator $\cK_{t,x_1, s_1,s_t} f(x) = \cL_{t, x_1,s_1} \varphi_f(x,s_t)$. Additionally, assume that the linear operator from \eqref{eq:joint-generator-linear} satisfies that $K \varphi_f(x,s) = K_x f(x)$ for some linear operator $K_x$, then the generator admits a linear parametrization of the form:
    \begin{align}
        \cK_t f(x) = \langle K_x f (x), F_t(x)\rangle\rev{, \qquad F_t(x) := \bE_{x_1,s_1,s_t}\left[F_{t,x_1,s_1}(x,s_t) \mid X_t = x\right].}
    \end{align}
\end{theorem}

\begin{figure}[t]
\begin{algorithm}[H]
\small
\caption{Generator Matching with an Auxiliary Process: recipe.}
\label{alg:gm-aux-recipe}
\begin{algorithmic}[0]
\State \textbf{Step 1.} Construct a conditional generator $\cL_{t,x_1,s_1}$ generating $p_{t|1}(x_t, s_t| x_1,s_1)$ and satisfying \eqref{eq:joint-generator-factorization}; \State \textbf{Step 2.} Parametrize $\cL_{t,x_1,s_1}$ through $F_{t,x_1,s_1}(x_t,s_t)$, this is the \emph{training target}.
\State \textbf{Step 3.} Parametrize $\cK_t$ (Theorem \ref{thm:joint-generators}) through $F_t^\theta(x)$
\State \textbf{Step 4.} Train $F_t^\theta$ by minimizing the loss $L_{\mathrm{cgm\text{-}aux}}(\theta)$ in \eqref{eq:conditional-generator-matching-aux}
\State \textbf{Step 5.} Sample by simulating the marginalized generator $\cK_t$ on $\bX$ driven by $F_t^\theta$.
\end{algorithmic}
\end{algorithm}
\vspace{-\baselineskip}
\vspace{-\baselineskip}
\end{figure}
This theorem specifies that given a generator $\cL_t$ generating $p_t(x,s)$ and acting on functions of $\bX \times \bS$,
one can define a new generator $\cK_t$ acting on functions over $\bX$ that generates the marginals $p_t(x)$. Furthermore, it  provides us with a principled condition in which the resulting generator can be linearly parametrized. A discussion on which processes satisfy such conditions, as well as a comparison to other related auxiliary-process constructions, is provided in Appendix~\ref{app:aux-gen-proofs}.

\textbf{The loss function:} Naively one could use the generator $\cK_t$ in the conditional generator matching objective in Equation \eqref{eq:conditional-generator-matching}. However, doing so would once again result in having to consider all the possible values of $S_t$. This would complicate the training as in \citet{campbell2023trans} and may be computationally intractable. To address this we use the following theorem:

\begin{theorem} (Conditional Generator Matching with Auxiliary Process) \label{thm:loss-function}
    \rev{Under the assumptions of Theorem~\ref{thm:joint-generators}, for any Bregman divergence $D$, possibly depending on $t$,} the loss given by
    \begin{align}\label{eq:conditional-generator-matching-aux}
        L_{\mathrm{cgm-aux}}(\theta)
        = \mathbb{E}_{t} \, \mathbb{E}_{x_1,s_1 \sim p_{\data}} \, \mathbb{E}_{x_t,s_t \sim p_{t|1}(\cdot \mid x_1,s_1)}
        \big[ D\!\left(F_{t,x_1,s_1}(x_t, s_t), F_t^\theta(x_t)\right) \big],
    \end{align}
    satisfies that $\nabla_\theta L_{\mathrm{gm}} = \nabla_\theta L_{\mathrm{cgm-aux}}$.
\end{theorem}

This theorem yields a tractable training objective and is central to our method. Evaluating the loss only requires sampling from the conditional probability paths of the joint process, which is often straightforward. Training on samples from the joint process---rather than the marginalized one---avoids explicitly summing over the many auxiliary configurations that can correspond to the same observed state. We summarize the framework for training with auxiliary processes in Algorithm~\ref{alg:gm-aux-recipe}.

\subsection{Designing a variable length generative model} \label{sec:practical-process}

We now instantiate step~1 of Algorithm~\ref{alg:gm-aux-recipe}: a conditional generator $\cL_{t,x_1,s_1}$ for variable-length data over an arbitrary state space $\tilde{\bV}$, which may itself be multimodal (e.g.\ $\tilde{\bV} = \bR^4 \times [N]$). We take an \emph{interpolant} perspective \citep{albergo2023stochastic}: we first describe what a sample $x_t$ looks like given a data point $x_1$, and then give the generator that produces it. The description has two steps, illustrated in Figure~\ref{fig:paper-overview}: every coordinate of $x_1$ receives an \emph{insertion time}, at which it is inserted with a value drawn from an \emph{insertion distribution}, and from then on it is \emph{denoised} toward its value.

\paragraph{The variable-length interpolant:} Let $x_1 \in \tilde\bV^{L}$. Fix an \emph{insertion distribution} $q(\cdot\mid v)$ on $\tilde\bV$, and a single-coordinate conditional probability path $p_u(\cdot\mid v)$, $u\in[0,1]$, from $q(\cdot\mid v)$ to $\delta_v$ with conditional generator $\cA_{u,v}$, as in standard Generator Matching. For example, one might have a Gaussian prior, $q(\cdot\mid v)=\pi$, with the flow matching path $u v+(1-u)\epsilon$. The insertion distribution then does not depend on $x_1$ and need not be learned. Alternatively, one may insert \emph{noisy data}: take $q(\cdot\mid v)$ to be the Gaussian path at an intermediate noise level and $p_u$ the same path from that level to the clean value. Now $q$ depends on $x_1$, and the model must additionally learn the insertion distribution; in the extreme case $q(\cdot\mid v)=\delta_v$ clean values are inserted directly and no denoising is needed.

\textit{(i) Insertion times.} Fix an increasing schedule $\gamma_t$ with $\gamma_0=0$ and $\gamma_1=1$. Each coordinate $i\in[L]$ receives an independent insertion time $\tau_i$ with $\bP(\tau_i>t)=1-\gamma_t$. The auxiliary variable is the ordered set of inserted coordinates $S_t=\{i:\tau_i\le t\}$, so that $|S_t|=\len(x_t)$ and $S_1=[L]$.

\textit{(ii) Insertion and denoising.} At time $\tau_i$, coordinate $i$ is inserted with a value drawn from $q(\cdot\mid x_1^{(i)})$. From then on it is denoised toward $x_1^{(i)}$ over the interval $[\tau_i,1]$, which we view as the coordinate's own \emph{denoising clock} $u(t;\tau) := (t-\tau)/(1-\tau)\in[0,1]$, running from $0$ at insertion to $1$ at $t=1$ irrespective of $\tau$. Hence $x_t^{(i)}\sim p_{u(t;\tau_i)}(\cdot\mid x_1^{(i)})$ independently for each inserted coordinate: a coordinate inserted late is simply denoised faster, and all are clean at the end. Since every coordinate is denoised on its own time, the construction resembles Diffuse Everything \citep{rojas2025diffuse}, where every modality has its own clock, and from which VL-DE takes its name.

The state is the sequence of inserted coordinates together with their insertion times, $x_t=((x_t^{(i)},\tau_i))_{i\in S_t}\in\bX:=\bigcup_{d\le D}(\tilde\bV\times[0,1])^d$. Insertion times are observed at sampling time and the denoiser needs them to know the noise level of each coordinate, so they belong to the data state rather than to the auxiliary variable.

\paragraph{The conditional generator:} The process above is Markov, and its generator has two terms mirroring the two steps:
\begin{align}\label{eq:full-forward-process}
    \cL_{t,x_1,s_1} = \Linsert_{t,x_1,s_1} + \Ldenoise_{t,x_1}.
\end{align}
Insertion is a jump process. Writing $s^{(k)}$ for the index in $x_1$ of the $k$-th entry of $x$ and $\ins_k(x,v,t)$ for the state obtained by inserting the entry $(v,t)$ at position $k$, its jump kernel is
\begin{align}\label{eq:insertion-kernel}
    R_t\bigl((x,S)\to(\ins_k(x,v,t),\,S\cup\{i\})\bigr)=\frac{\dot\gamma_t}{1-\gamma_t}\,\indicator\{s^{(k-1)}<i<s^{(k)}\}\,q\bigl(v\mid x_1^{(i)}\bigr),
\end{align}
with the conventions $s^{(0)}=0$ and $s^{(\len(x)+1)}=L+1$; it inserts in between existing coordinates as in \citet{kim2025any}. The kernel has a \emph{rate} part, which depends on the state only through $S$, and an \emph{initialization} part $q$, which is known in closed form when it is a prior and must be learned otherwise. Denoising runs the base generator of every inserted coordinate on its clock, whose speed is $\partial_t u(t;\tau)=1/(1-\tau)$:
\begin{align}\label{eq:denoise-generator}
    \Ldenoise_{t,x_1}f(x,s)=\sum_{k=1}^{\len(x)}\frac{1}{1-\tau^{(k)}}\;\cA^{(k)}_{u(t;\tau^{(k)}),\,x_1^{(s^{(k)})}}f(x,s),
\end{align}
with $\cA^{(k)}$ acting on the $k$-th value only. Since the rate depends on the state only through $S$ and denoising never changes $S$, the two steps decouple:

\begin{theorem}\label{thm:conditional-prob-path-factorization}
\rev{The process defined by steps (i)--(ii) is Markov with generator \eqref{eq:full-forward-process}, and its} conditional probability path factorizes as
\begin{align}\label{eq:conditional-prob-path-factorization}
    p_t(x_t,S_t\mid x_1,S_1)=p_t(S_t,\tau\mid S_1)\prod_{i\in S_t}p_{u(t;\tau_i)}\bigl(x_t^{(i)}\mid x_1^{(i)}\bigr),
\end{align}
where $\tau=(\tau_i)_{i\in S_t}$ are the insertion times contained in $x_t$. In particular, $(x_t,S_t)$ is sampled by steps (i)--(ii) above, without simulating the process.
\end{theorem}

\paragraph{The loss function:} By Theorem~\ref{thm:joint-generators}, at sampling time the model observes only $x_t$, i.e.\ values and insertion times, and outputs the denoising parameters of every coordinate, the expected number of missing coordinates in each gap $j$ between consecutive entries, and, when $q$ is not a prior, a distribution $q^\theta_t(\cdot\mid x_t,j)$ for the value of the next coordinate inserted in gap $j$. Theorem~\ref{thm:loss-function} then yields the objective.

\begin{lemma}\label{lemma:loss-for-our-case}
For the generator in \eqref{eq:full-forward-process}, the training objective is
\begin{align*}
\bE\Big[
\underbrace{D\!\left(F^{\text{denoise}}_{t,x_1,s_1}(x_t,s_t),F^{\theta,\text{denoise}}_t(x_t)\right)}_{\text{denoise}}
&+
\underbrace{\sum_{j}\phi\!\left(s_t[j]-s_t[j-1]-1,\,F^{\theta,\text{insert}}_t(x_t,j)\right)}_{\text{insert}}
\Big.\\
\Big.
&+
\underbrace{\sum_{j}\ \sum_{s_t[j-1]<i<s_t[j]}\bE_{v\sim q(\cdot\mid x_1^{(i)})}\!\left[-\log q^\theta_t(v\mid x_t,j)\right]}_{\text{initialization}}
\Big],
\end{align*}
where $t\sim\mathcal U[0,1]$, $x_1\sim\pdata$ and $(x_t,s_t)\sim p_{t\mid 1}(\cdot\mid x_1,s_1)$ as in Theorem~\ref{thm:conditional-prob-path-factorization}. The $k$-th entry of the denoising target is the base target of $\cA$ at clock time $u(t;\tau^{(k)})$ toward $x_1^{(s^{(k)})}$, and $\phi(x,y)=y-x\log y$ is the standard Bregman divergence for jump rates \citep{kim2025any}. When $q$ is a prior, the initialization term is constant in $\theta$ and is dropped. \rev{Other positive time reweightings of the terms are possible (Remark~\ref{rem:reweighting}).} See Appendix~\ref{app:process-proofs} for a proof.
\end{lemma}
\noindent
All targets are read off from the sampled insertion times, with no marginalization over the latent insertion orders that could have produced $x_t$. The initialization term is the cross-entropy of the learned insertion distribution against the true one; it plays the role of the unmasking loss in masked diffusion and is only needed when coordinates are inserted as noisy data.

\paragraph{Extensions:} The recipe is modular. Other insertion kernels, e.g.\ auto-regressive or block insertion, may replace \eqref{eq:insertion-kernel} (Appendix~\ref{app:insertion-methods}). Inserting mask tokens recovers FlexMDM \citep{kim2025any}, and inserting clean values with substitutions recovers Edit Flows \citep{havasi2025edit}. \emph{Deletions} follow by augmenting $x_1$ with junk coordinates whose lifetime ends before $t=1$, and Theorem~\ref{thm:conditional-prob-path-factorization} and Lemma~\ref{lemma:loss-for-our-case} extend naturally (Appendix~\ref{app:deletions}). Finally, $\tilde\bV$ may itself be a variable-length space, so modalities with \emph{independently varying lengths} are handled with one clock per modality as in Diffuse Everything, unlike previous approaches which require equal lengths.\label{sec:mixed-length-datasets}

\subsection{Higher-Order Sampling via Autonomized Operator Splitting} \label{sec:higher-order-samplers}
Higher-order integrators can substantially reduce the number of sampling steps in diffusion models \cite[e.g.,][]{karras2022elucidating,zhanggddim,lu2022dpm,ren2025fast}. In multimodal or variable-length settings, however, operations like insertion alter the dynamics and make standard higher-order samplers hard to apply directly. Existing methods therefore often rely on Euler discretizations with many steps (e.g., \cite{billera2025branching} uses $1000$).

To address this, we design a sampler based on \emph{autonomized operator splitting}. While motivated by our framework, the method applies more generally to time-dependent Markov generators, so we present the method in its full generality. Consider a time-dependent generator of the form $
    \cK_t = \cK_t^{A} + \cK_t^{B}$.

If there were no time dependence, high-order integration could be accomplished via operator splitting~\citep{mclachlan2002splitting} based on $\exp(\cK^{A,*} h)$ and $\exp(\cK^{B,*} h)$. However, the non-autonomy, or more precisely, the lack of commutativity in time ($[\cK_{t_1}^{A}, \cK_{t_2}^{A}] \neq 0$, $[\cK_{t_1}^{B}, \cK_{t_2}^{B}] \neq 0$), is what makes accurate numerical integration nontrivial as the exact flow maps induced by $\cK^{A,*}_t$ and $\cK^{B,*}_t$ are no longer tractable. We address this by introducing auxiliary time variables $t_1(t)$ and $t_2(t)$ in addition to $p_t$ and split the dynamics in a very specific way:
\begin{align}\label{eq:operator-decomposition}
    \partial_t 
    \begin{pmatrix}
        p_t(x) \\
        t_1 \\
        t_2
    \end{pmatrix}
    =
    \begin{pmatrix}
        \left(\cK_{t_1}^{A,*}+\cK_{t_2}^{B,*}\right) p_t(x)\\
        1\\
        1
    \end{pmatrix}
    =
    \begin{pmatrix}
        \cK_{t_1}^{A,*} p_t(x)\\
        0\\
        1
    \end{pmatrix}
    +
    \begin{pmatrix}
        \cK_{t_2}^{B,*} p_t(x)\\
        1\\
        0
    \end{pmatrix}.
\end{align}
Note the split systems $    \partial_t 
    \begin{pmatrix}
        p_t(x) \\
        t_1 \\
        t_2
    \end{pmatrix}
    =     \begin{pmatrix}
        \cK_{t_1}^{A,*} p_t(x)\\
        0\\
        1
    \end{pmatrix}$ and $    \partial_t 
    \begin{pmatrix}
        p_t(x) \\
        t_1 \\
        t_2
    \end{pmatrix}
    =     \begin{pmatrix}
        \cK_{t_2}^{B,*} p_t(x)\\
        1\\
        0
    \end{pmatrix}$
    both became autonomous system because $t_1$ is a constant in A dynamics and $t_2$ is a constant in B dynamics, and their analytical solutions are respectively given by the following flow maps:
\begin{align}
    \varphi_A^h(p_t, t_1, t_2)
    =
    \begin{pmatrix}
        \exp(\cK_{t_1}^{A,*} h) p_t(x) \\
        t_1 \\
        t_2 + h
    \end{pmatrix},
    \qquad
    \varphi_B^h(p_t, t_1, t_2)
    =
    \begin{pmatrix}
        \exp(\cK_{t_2}^{B,*} h) p_t(x) \\
        t_1 + h \\
        t_2
    \end{pmatrix}.
\end{align}

\paragraph{Second-order sampling.}
Applying Strang splitting yields the second-order scheme
\begin{align}
    X_{t+h} \sim 
    \varphi_A^{h/2}
    \circ
    \varphi_B^{h}
    \circ
    \varphi_A^{h/2}
    \,\delta_{(X_t,t_1,t_2)}.
\end{align}

Naively, this requires three neural network evaluations per step. However, consecutive steps contain adjacent $\varphi_A^{h/2}$ operations, which can be merged using the semigroup property:
\begin{align}
    \varphi_A^{h/2}
    \circ
    \varphi_B^{h}
    \circ
    \underbrace{\varphi_A^{h/2} \circ \varphi_A^{h/2}}_{\varphi_A^h}
    \circ
    \varphi_B^{h}
    \circ
    \underbrace{\varphi_A^{h/2} \circ \varphi_A^{h/2}}_{\varphi_A^h}
    \circ
    \varphi_B^{h}
    \cdots
\end{align}
This reduces the cost from three to two neural network evaluations per step. Pseudocode for both variants is given in Listings~\ref{code:staggered} and~\ref{code:split} in Appendix~\ref{app:sampler-pseudocode}.

\paragraph{Applying this to variable length generation:} In our setting we set $\cK_t^A$ to be the denoising operator, while $\cK_t^B$ corresponds to the insertion operator. This decouples the insertion of new elements from their refinement. In mixed-length multimodal settings we follow this same convention. 
\vspace{-\baselineskip}
\section{Experiments}
\vspace{-.7\baselineskip}
We evaluate VL-DE on layout generation, where each element is a discrete category and a continuous bounding box and the number of elements varies. Prior layout models use heuristics for both of these challenges, such as quantization of the coordinates or embedding the categories into the Euclidean space~\citep{zhang_layoutdiffusion,guerreiro2024layoutflow}, and receive the number of elements as input or pad to a fixed number of slots. On the other hand, VL-DE generates the length and keeps each modality in its native space. We evaluate it unconditionally on RICO~\citep{deka2017rico} and PubLayNet~\citep{zhong2019publaynet}, and on CGL~\citep{zhou2022cglgan}, where the length must be inferred from the canvas; a demonstration where each box includes variable length text is included in Appendix~\ref{app:crello-details}, other experimental details are in Appendix~\ref{app:hyperparameters}. \rev{We further evaluate on 3D molecule generation on QM9 (Section~\ref{sec:qm9-experiments}), and a new toy problem of arithmetic equations,(Section~\ref{sec:arith-experiments}).}
\begin{table}[b]
    \centering
    \caption{\textbf{Unconditional layout generation.} \textbf{Left:} layout models grouped by how they handle the number of elements (numbers from \citet{guerreiro2024layoutflow}; $*$ retrained there). \textbf{Right:} methods that generate the number of elements, on RICO, grouped by how they represent boxes. $\rightarrow$: closer to the validation data is better. Protocol and sampling steps in Appendix~\ref{app:layout-details}.}
    \label{tab:layout-results}
    \scriptsize
    \setlength{\tabcolsep}{3pt}
    \renewcommand{\arraystretch}{0.9}
    \setlength{\aboverulesep}{0.2ex}\setlength{\belowrulesep}{0.3ex}
    \begin{minipage}[t]{0.615\linewidth}
    \centering
    \begin{tabular}{lrrrrrrr}
        \toprule
        & \multicolumn{3}{c}{RICO} && \multicolumn{3}{c}{PubLayNet}\\
        \cmidrule{2-4}\cmidrule{6-8}
        Model & \abbfid & \abbalignment & \abboverlap && \abbfid & \abbalignment & \abboverlap \\
        \midrule
        \multicolumn{8}{l}{\textit{Fixed number of slots, padded with an empty class}}\\
        LayoutDM$^*$      & 4.43  & 0.143 & 0.584 && 36.85 & 0.180 & 0.132 \\
        LayoutDiffusion   & 2.49  & 0.069 & 0.502 && 8.63  & 0.065 & 0.003 \\
        \midrule
        \multicolumn{8}{l}{\textit{Number of elements given as input}}\\
        DLT$^*$           & 13.02 & 0.271 & 0.571 && 12.70 & 0.117 & 0.036 \\
        LayoutFlow        & 2.37  & 0.150 & 0.498 && 8.87  & 0.057 & 0.009 \\
        \midrule
        \multicolumn{8}{l}{\textit{Number of elements generated}}\\
        \rowcolor{myhighlight} VL-DE (ours)              & 2.32  & 0.253 & 0.547 && \rev{8.45}  & \rev{0.130} & \rev{0.014} \\
        \midrule
        Validation data                                  & 2.10  & 0.093 & 0.466 && 8.10  & 0.022 & 0.003 \\
        \bottomrule
    \end{tabular}
    \end{minipage}\hfill
    \begin{minipage}[t]{0.365\linewidth}
    \centering
    \begin{tabular}{lrrr}
        \toprule
        & \multicolumn{3}{c}{RICO}\\
        \cmidrule{2-4}
        Method & \abbfid & \abbalignment & \abboverlap \\
        \midrule
        \multicolumn{4}{l}{\textit{Discrete: boxes quantized to tokens}}\\
        Edit Flows        & 6.43 & 0.191 & 0.570 \\
        FlexMDM           & 8.41 & 0.188 & 0.606 \\
        \midrule
        \multicolumn{4}{l}{\textit{Multimodal: continuous boxes}}\\
        OneFlow           & 3.31 & 0.293 & 0.499 \\
        Branching Flows   & 12.09 & 0.247 & 0.643 \\
        \midrule
        \multicolumn{4}{l}{\textit{Multimodal, arbitrary state space}}\\
        \rowcolor{myhighlight} VL-DE (ours)              & 2.32 & 0.253 & 0.547 \\
        \midrule
        Validation data                                  & 2.10 & 0.093 & 0.466 \\
        \bottomrule
    \end{tabular}
    \end{minipage}
\end{table}

\subsection{Layout Generation on RICO and PubLayNet} \label{sec:layout-experiments}
We set $\tilde\bV=[C]\times[0,1]^4$ with a mask prior for the category and Gaussian noise as the box prior (Figure~\ref{fig:paper-overview}), and use an insertion schedule that only inserts elements during $t\in[0,0.5]$; ablations of both choices can be found in Appendix~\ref{app:rico-insertion}.

\textbf{Setup.} We follow the training and evaluation protocol of LayoutFlow~\citep{guerreiro2024layoutflow} on RICO~\citep{deka2017rico}, $66$K mobile UI layouts with $25$ categories, and PubLayNet~\citep{zhong2019publaynet}, $360$K document layouts with $5$ categories. We report FID in the feature space of \citet{zhang_layoutdiffusion}, together with the geometric statistics Alignment and Overlap, whose ideal value is that of the validation data. The standard conditional tasks, generation given the categories (\tasktype) or the categories and sizes (\tasktypesize), completion and refinement, are generation modes of Section~\ref{sec:practical-process} and need no task-specific training. Other details can be found in Appendix~\ref{app:layout-details}.

\textbf{VL-DE matches fixed-length approaches.} Table~\ref{tab:layout-results} (left) compares VL-DE with the layout generation literature on RICO and PubLayNet. VL-DE neither receives the length nor pads, yet it matches the best of these models on RICO, reaching a new state-of-the-art FID of $2.32$ against $2.37$ for LayoutFlow. Unselected samples of every method are shown in Figure~\ref{fig:layout-qualitative} (Appendix~\ref{app:layout-details}).

\textbf{Comparison against other variable-length methods.} A natural question is how existing variable-length approaches perform on this task. We test this by running FlexMDM and Edit Flows on the tokenized problem of \citet{chai2023layoutdm}, and OneFlow and Branching Flows on the multimodal setup; details are in Appendix~\ref{app:layout-details}. Table~\ref{tab:layout-results} (right) shows that VL-DE outperforms all of them, with OneFlow close behind. We attribute this to the native multimodal dynamics of our method, in which each box evolves jointly with its category rather than separately as in OneFlow. Branching Flows severely underperforms here, as its duplication dynamics may not be well suited to layout generation.

\begin{figure}[t]
    \centering
    \includegraphics[width=\linewidth]{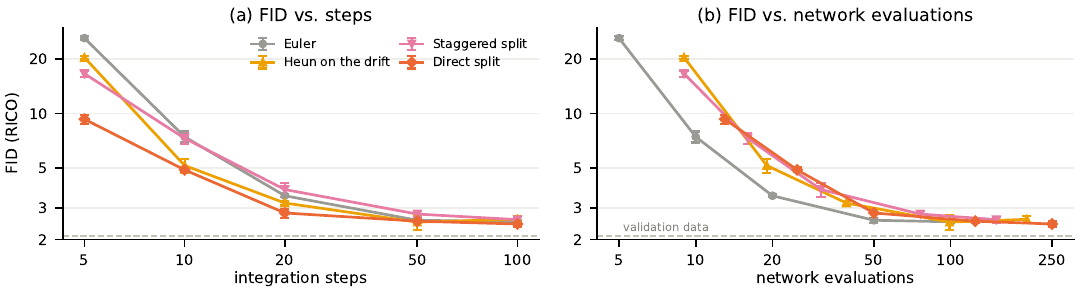}
    \caption{\textbf{Operator-splitting samplers on RICO} (Mean over $5$ sampling seeds, error bars one standard deviation). (a) Per step, the direct split is far ahead of Euler below $20$ steps. (b) Per network evaluation, Euler is the most economical sampler at every budget.}
    \label{fig:rico-samplers}
    \vspace{-\baselineskip}
\end{figure}

\textbf{Our samplers improve generation at low steps:} Figure~\ref{fig:rico-samplers} compares the samplers of Section~\ref{sec:higher-order-samplers} on the learned-initialization model for $5$ to $100$ steps. Per step, the direct split reaches an FID of $9.3$ at $5$ steps and $4.9$ at $10$, against $26.2$ and $7.4$ for Euler, and the two match from $50$ steps on. The staggered variant does not retain this gain. Per NFE, Euler is best at every budget, so the splitting samplers pay off when the number of steps rather than the number of evaluations is the bottleneck (\rev{Convergence analysis in (Appendix~\ref{app:sampler-order})).} We report all other results with Euler for a fair comparison.

\subsection{Content-Aware Layout Generation on CGL} \label{sec:cgl-experiments}
Unconditionally, the length distribution can be estimated from the data itself; the conditional setup does not, for instance, there may be a single example per condition, which makes the length much harder to model. We study this on content-aware layout generation, where the input is a poster canvas and the output its layout. Our goal is not the best content-aware generator but a comparison of how length can be modeled: sampling it from the training histogram, which ignores the canvas; \emph{padding} to a fixed number of slots with an empty class~\citep{zhou2022cglgan,li2026layoutdit}; \emph{autoregressive} decoding with an end token~\citep{horita2024ralf}; and our \emph{insertion} based approach.

\textbf{Setup.} We use CGL~\citep{zhou2022cglgan} with the splits, inpainted canvases, saliency maps and evaluator released by RALF~\citep{horita2024ralf}, and report their metrics on the test split: FID, saliency occlusion (Occ), strict underlay effectiveness (Und$_s$) and overlay (Ove) in Table~\ref{tab:cgl-results}, with unreadability and loose underlay in Appendix~\ref{app:cgl-details}. Experimental details are described in Appendix~\ref{app:cgl-details}.

\begin{figure}[t]
    \centering
    \begin{minipage}[c]{0.6\linewidth}
    \centering
    \footnotesize
    \setlength{\tabcolsep}{4pt}
    \resizebox{\linewidth}{!}{%
    \begin{tabular}{lrrrr}
        \toprule
        Length mechanism & \abbfid & Occ$\downarrow$ & Und$_s\uparrow$ & Ove$\downarrow$ \\
        \midrule
        Histogram (canvas-blind)        & 1.63 & 0.318 & 0.877 & 0.006  \\
        Padding + empty class           & 4.02 & 0.144 & 0.774 & 0.013  \\
        Autoregressive (RALF w/o retr.) & 2.89 & 0.125 & 0.917 & 0.011  \\
        \rowcolor{myhighlight} Insertion (VL-DE, ours) & 1.53 & 0.148 & 0.950 & 0.0097 \\
        \midrule
        RALF + retrieval~\citep{horita2024ralf} & 1.32 & 0.126 & 0.978 & 0.0042 \\
        Test data                                & 0.80 & 0.125 & 0.988 & 0.0003 \\
        \bottomrule
    \end{tabular}}
    \captionof{table}{\textbf{Content-aware layout generation on CGL} VL-DE outperforms other base methods in conditional generation. Full metrics, in Table~\ref{tab:cgl-results-full}; samples in Figure~\ref{fig:cgl-qualitative}.}
    \label{tab:cgl-results}
    \end{minipage}\hfill
    \begin{minipage}[c]{0.38\linewidth}
    \centering
    \includegraphics[width=\linewidth]{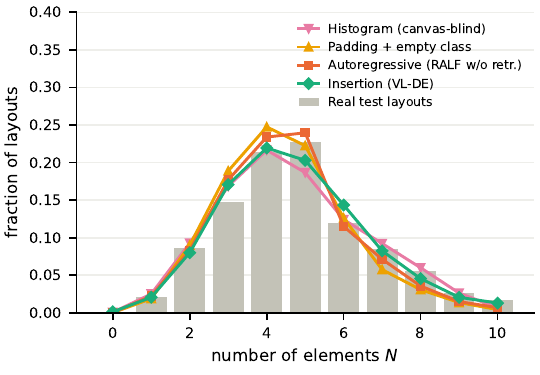}
    \caption{\textbf{Generated count distributions on CGL} VL-DE better models the length distribution.}
    \label{fig:cgl-counts}
    \end{minipage}
    \vspace{-\baselineskip}
\end{figure}

\textbf{VL-DE better models the length and the sample distribution.} Table~\ref{tab:cgl-results} shows the results for the given task, among the existing methods VL-DE obtains the lowest FID and ranks highly in other metrics. Furthermore, Figure~\ref{fig:cgl-counts} shows the histograms for each method where we observe that VL-DE is the best at modeling the length, proving the robustness of our method on the conditional setting.

\begin{revblock}
\subsection{Molecule Generation on QM9} \label{sec:qm9-experiments}
Molecules are a variable-length set of atoms, with a 3D position and a discrete element type. On QM9~\citep{ramakrishnan2014quantum} we set $\tilde\bV=\bR^3\times[5]$: the 3D positions are inserted from a Gaussian distribution, while the atom type is drawn from a learned insertion distribution $q^\theta$ which is not denoised further. We use the same parametrization as in \cite{campbell2023trans,satorras2021n}  and the same insertion schedule as in the layouts experiments.

\textbf{Setup.} We follow the protocol of Branching Flows~\citep{billera2025branching}: $10$K generated molecules are compared with $10$K reference molecules using the Kolmogorov Smirnov statistic (KS) for several distribution metrics and a binned multivariate Jensen--Shannon divergence (JSDm) over their joint distribution. For sampling we use $1000$ Euler steps, as in their evaluation. Details are in Appendix~\ref{app:qm9-details}.

\textbf{Results.} Table~\ref{tab:qm9-results} shows that VL-DE is on par with Branching Flows and the fixed-length ADiT~\citep{joshi2025all}, which is given the number of atoms, and is far ahead of Transdimensional Jumps~\citep{campbell2023trans}. Demonstrating that VL-DE extends naturally to other domains.
\end{revblock}

\begin{table}[t]
    \begin{revblock}
    \centering
    \caption{\textbf{Unconditional molecule generation on QM9.} Multivariate Jensen--Shannon divergence (JSDm, lower is better), $1-$KS distance to the data (higher is better) and fractions of valid and unique molecules; $10$K samples. Baselines from \citet{billera2025branching};}
    \label{tab:qm9-results}
    \scriptsize
    \setlength{\tabcolsep}{3pt}
    \renewcommand{\arraystretch}{0.9}
    \resizebox{\linewidth}{!}{%
    \begin{tabular}{lrrrrrrrrrrrrrr}
        \toprule
        Model & JSDm$\downarrow$ & \#Atoms & HBD & HBA & LogP & Weight & \#C & \#H & \#O & \#N & \#F & Valid & Unique & Avg.\ $1-$KS \\
        \midrule
        \multicolumn{15}{l}{\textit{Models with length access}}\\
        \midrule
        ADiT & 0.086 & 0.988 & 0.990 & 0.923 & 0.976 & 0.944 & 0.978 & 0.989 & 0.945 & 0.973 & 0.999 & 0.985 & 0.979 & 0.971 \\
        Multimodal FM & 0.063 & 0.993 & 0.981 & 0.964 & 0.967 & 0.931 & 0.988 & 0.985 & 0.988 & 0.980 & 0.996 & 0.989 & 0.965 & 0.977 \\
        \midrule
        \multicolumn{15}{l}{\textit{Variable length models}}\\
        \midrule
        Transdimensional Jumps & 0.120 & 0.870 & 0.866 & 0.832 & 0.833 & 0.883 & 0.815 & 0.856 & 0.925 & 0.855 & 0.968 & 0.978 & 0.986 & 0.870 \\
        Branching Flows & \textbf{0.080} & \textbf{0.990} & 0.982 & 0.948 & 0.952 & 0.902 & 0.984 & \textbf{0.979} & 0.981 & 0.978 & 0.997 & \textbf{0.987} & 0.973 & 0.969 \\
        \rowcolor{myhighlight} VL-DE (ours) & 0.082 & 0.943 & \textbf{0.987} & \textbf{0.987} & \textbf{0.989} & \textbf{0.918} & \textbf{0.987} & 0.950 & \textbf{0.993} & \textbf{0.996} & \textbf{0.999} & 0.970 & \textbf{0.979} & \textbf{0.975} \\
        \bottomrule
    \end{tabular}}
    \end{revblock}
    \vspace{-\baselineskip}
\end{table}

\begin{figure}[t]
    \begin{revblock}
    \centering
    \includegraphics[width=\linewidth]{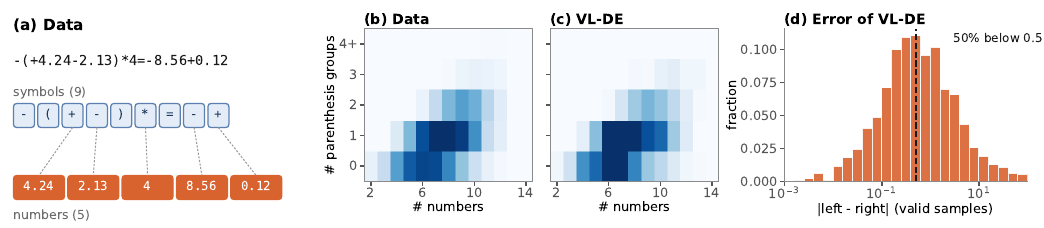}
    \caption{\textbf{Arithmetic equations.} (a) The symbols and numbers of an equation; the number of symbols is set by the numbers and the parenthesis groups. (b, c) Joint distribution of these counts for the test data and $1000$ samples of VL-DE. (d) $|$left $-$ right$|$ of the valid samples of VL-DE.}
    \label{fig:arith-overview}
    \end{revblock}
    \vspace{-\baselineskip}
\end{figure}

\begin{revblock}
\subsection{Arithmetic Equations: Modalities of Different Lengths} \label{sec:arith-experiments}
We introduce a new toy problem of arithmetic equations Figure~\ref{fig:arith-overview}, it consists of a sequence of symbols and a sequence of numbers that satisfy an equality condition, each with different length. Under this setup the symbols and numbers are separate sequences that have a coupling induced by the data, and the goal is to generate both of them without leveraging extra knowledge. 

\textbf{Setup.} We generate $2$M training equations from random arithmetic operations whose two sides evaluate to the same value; the generation procedure, model and training details are in Appendix~\ref{app:arith-details}. We sample $1000$ equations with $800$ Euler steps and measure whether their lengths are consistent with the mathematical rules and the discrepancy between both sides of the equality.

\textbf{Results.} VL-DE closely matches the joint distribution of the number of numbers and of parenthesis groups (Figure~\ref{fig:arith-overview}b,c), demonstrating that VL-DE can learn the coupling between lengths. Furthermore, half of the generated equations have a discrepancy smaller than $0.5$. Considering that small differences in products can result in big discrepancies, this is proves that VL-DE can learn the true distribution (Figure~\ref{fig:arith-overview}d). We also present an ablation against an interleaved model in Appendix~\ref{app:arith-details}.
\end{revblock}

\section{Limitations and Future Work}
\vspace{-.5\baselineskip}
We highlight a few limitations of the current work. First, while the framework supports arbitrarily many modalities, all of our experiments use two; scaling to substantially more modalities is an interesting direction. Beyond these, VL-DE points to several broader directions: richer auxiliary-process designs (e.g., kernels that mix insertion with reordering or substitutions), and applying the autonomized operator-splitting sampler to non-autonomous generative models beyond the variable-length setting. On layouts, the sampler's gain per step does not carry over to a gain per network evaluation; settings where one operator is cheap to evaluate, such as a separate insertion network, or caching mechanisms, could turn it into a significant gain in cost.

\newpage
\bibliography{iclr2027_conference}
\bibliographystyle{plainnat}

\newpage
\appendix
\ifdefined\arxivversion\else 
\subsection*{AI use statement}

Generative AI was utilized for verifying proofs, formulating mathematical claims, assisting in the writing of proofs, suggesting edits to mathematical formulations, and designing or providing feedback on research methodology and experiments. All theorems and proofs originated as constructs by the authors, with AI-suggested edits and changes added after the fact. Proofs are verified as correct by the authors and also independently verified by AI checks. 

For the experimental code, the authors first wrote a base implementation of the estimators, samplers, and evaluation pipeline. Generative AI was subsequently used to write the scripts that launch and organize experimental runs (e.g., sweeps over hyperparameters, seeds, and problem dimensions), to revise the existing code base, and to produce the plotting code. All AI-generated or AI-modified code was reviewed by the authors. Generative AI was also used as a writing aid, offering edits to improve the clarity and readability of the text.
\fi

\section{Diffuse Everything within Generator Matching} \label{app:de-gm}
Diffuse Everything was formulated in the Denoising Markov Models framework~\citep{benton2024denoising,ren2025unified}, which only admits generative processes that reverse a forward noising process. Generator Matching~\citep{holderrieth2024generator} imposes no such requirement, so it covers a strictly larger class of generators, and the construction of Diffuse Everything carries over to it unchanged: each modality receives its own generator and its own time variable, and the joint generator is their sum.

The decoupled time variables only change how training times are sampled. Drawing a separate time $(t_1,t_2)$ per modality on paired data trains a single model for both unconditional and conditional generation, since any modality can be held at $t=1$ at sampling time. Drawing a shared time $t_1=t_2$ gives up conditional generation but trains the model only on the diagonal it uses for unconditional sampling. Unconditionally, a model trained with a shared time therefore behaves like one trained directly with Generator Matching, whereas decoupled times trade some unconditional quality for the additional generation modes. 

\section{Related Works}

We organize related work into unimodal variable-length methods, interleaving-based multimodal methods, and general-state-space multimodal methods, and discuss how each relates to VL-DE. Table~\ref{tab:method_comparison} summarizes the comparison.

\textbf{Branching Flows} The work most closely related to ours is Branching Flows \citet{billera2025branching}. They propose a general framework, similar in spirit to ours, for modeling variable-length data. Unlike our approach, variable length is handled through explicit splits during the generation process. Each resulting branch is then denoised independently until the data distribution is recovered.
A key advantage of this strategy is that it avoids the need to explicitly model the distribution of marginals at intermediate times 
t. However, this comes at the cost of requiring more aggressive velocity fields, which can become problematic when neighboring elements of the data distribution differ significantly in magnitude. In contrast, this approach performs particularly well in settings where neighboring points are close to each other, such as molecule generation.
Like our method, Branching Flows are agnostic to the data modality (e.g., discrete, Euclidean, or manifold-valued data) and can be paired with arbitrary flow models.

\textbf{Flex MDM} Another relevant work that inspired our approach is Flex MDM \cite{kim2025any}, which addresses variable-length discrete data using mask diffusion models. Their method introduces masked token insertions, which are subsequently unmasked following the standard mask diffusion framework. Their approach corresponds to a special case of our setting when $\Ldenoise$ is trivial and random insertions are performed on discrete data. From this perspective, our method recovers the benefits of FlexMDM while extending it to other data modalities as well as more general classes of flow models.

\textbf{Transdimensional Diffusion} \cite{campbell2023trans} is one of the earlier works on modeling variable length data using diffusion models. It models variable length euclidean data by using jump diffusions. A forward process is designed to delete dimensions, while the backwards process is learned to insert dimensions.
A key difference from our approach is that this method does not leverage an auxiliary process, resulting in an objective that involves a complex mixture over the data distribution. This significantly complicates training and, in practice, necessitates simplifying assumptions that introduce discrepancies between the theoretical objective and the implemented training procedure.
Moreover, insertions are performed directly in the data space rather than via mask tokens, as in our method. While this avoids the need for a separate insertion head, it further increases the complexity of the learning objective. In this sense, both our approach and Branching Flows benefit from a simpler and more structured treatment of insertions.

\textbf{Edit Flows} \cite{havasi2025edit} propose another method for modeling variable length in discrete spaces. They combine substitutions, insertions and deletions to model the data distribution. The combination of insertions and deletions allows modeling the length in a flexible manner. Inserted elements follow the marginal distribution at a given time and can subsequently be refined through substitutions, similar in spirit to our denoising of inserted coordinates.

\textbf{OneFlow} \citet{nguyen2025oneflow} extends the Edit Flows framework to Euclidean data by inserting an image token equipped with its own time variable; after insertion, a noise variable associated with the token is progressively denoised. While this extension enables certain multimodal settings, it imposes structural constraints: the modalities must be paired one-to-one within a single sequence, and the data is restricted to a sequential insertion-and-refinement composition.

\textbf{Transfusion} \cite{zhoutransfusion} interleaves discrete tokens and continuous values into a single sequence and applies a unified transformer that handles each position with the appropriate per-token loss (cross-entropy for discrete, denoising for continuous). Like One Flow, it is tied to the specific pairing of one discrete and one continuous modality and does not extend naturally to general state spaces or to more than two modalities. Its variable-length behaviour is also inherited from the underlying autoregressive backbone rather than by defining a stochastic process as the other approaches considered here.

\begin{table}[t]
    \centering
    \caption{Comparison of methods for modeling variable-length data. \emph{Inserted object}: what is added when the dimensionality grows. \emph{Modalities}: what state spaces are supported. \emph{Arbitrary denoiser}: whether the per-coordinate dynamics can be any generator on $\tilde\bV$ (as opposed to a fixed mask-diffusion or noisy-data scheme). \emph{Different per-modality length}: whether each modality can carry its own independent length.}
    \label{tab:method_comparison}
    \setlength{\tabcolsep}{1.6pt}
    \renewcommand{\arraystretch}{1.0}
    \begin{tabular}{
        p{2.8cm}
        >{\centering\arraybackslash}p{2.6cm}
        >{\centering\arraybackslash}p{2.6cm}
        >{\centering\arraybackslash}p{2.6cm}
        >{\centering\arraybackslash}p{2.6cm}
        >{\centering\arraybackslash}p{2.8cm}
    }
        \toprule
        \textbf{Method}
        & \textbf{Inserted Object}
        & \textbf{Modalities}
        & \textbf{Arbitrary denoiser}
        & \textbf{Different per-modality length} \\
        \midrule
        Flex MDM
        & Mask tokens
        & Discrete
        & \texttimes
        & \texttimes \\

        Transdimensional Diffusion
        & Noisy data
        & Euclidean
        & \texttimes
        & \texttimes \\
        
        Edit Flows
        & Noisy data
        & Discrete
        & \texttimes
        & \texttimes \\

        OneFlow
        & Noisy data
        & Euclidean + Discrete
        & \texttimes
        & \texttimes \\

        Transfusion
        & Clean Tokens / Noisy Images
        & Euclidean + Discrete
        & \texttimes
        & \texttimes \\

        Branching Flows
        & Duplicates
        & General
        & \checkmark
        & \texttimes \\

        \midrule
        \textbf{VL-DE (Ours)}
        & Arbitrary
        & General
        & \checkmark
        & \checkmark \\
        \bottomrule
    \end{tabular}
    \label{tab:comparisonMethods}
\end{table}

\section{Background and Proofs}
\subsection{Background on Generators}\label{app:background-generators}
In this section we provide some definitions regarding generators of Markov processes.
\begin{definition} (Generator of a process)
Given a time homogenous process $X_t$ we can define the generator as:
\begin{align}\label{eq:def-generator}
    \cL f &= \lim_{h\to 0+} \frac{1}{h} \bE[f(X_h) - f(X_0) | X_0 = x] 
\end{align}
for $f$ in the domain of $\cL$.
\end{definition}
In practice it is common to consider processes that are not time homogenous, for this reason it is useful to consider an augmented process $\tilde{x}_t = (x_t, t)$, this process also has an augmented generator $\tilde{\cL}$ that satisfies the relation:
\begin{align*}
    \tilde{\cL}_t f(\tilde{x}) = \cL f(x,t) = \partial_t f(x,t) + \cL_t f(x,t)
\end{align*}
where $\displaystyle \cL_t f(x,t) = \lim_{h\to 0+} \frac{1}{h} \bE[f(X_{t+h}, t) - f(X_t, t) | X_t = x] $ is the generator at time $t$. A notable property of generators is that it allows us to define an equation for the time evolution of the density of our process, to do so we need the notion of the adjoint operator.
\begin{definition} (Adjoint Operator)
We define the adjoint operator $\cL^*$ with respect to the inner product $\langle\cdot, \cdot \rangle$ on $L^2(E, \mu)$ as the unique operator satisfying:
\begin{align*}
    \langle \cL f, g \rangle = \langle f , \cL^* g \rangle
\end{align*}
\end{definition}

Given the adjoint operator we can write down the Kolmogorov Forward Equation, otherwise known as Fokker-Planck equation. 
\begin{theorem} (Fokker-Planck Equation)
The marginal density $p_t(x)$ of a process with generator $\cL$ satisfies:
\begin{align*}
    \partial_t p_t = \cL^*_t p_t
\end{align*}
Alternatively, in its weak form the equation above becomes:
\begin{align}\label{eq:fokker-planck-weak}
    \partial_t \int f(x) p_t(x) = \int \cL_t f(x)  p_t(x)
\end{align}
for all test functions $f$. For a proof refer to Theorem A.23 of \citet{ren2025unified}.
\end{theorem}

For \textbf{jump components} of the process, we specify the dynamics through a jump kernel $R_t(z,z' \mid x_1)$, which gives the transition rate from state $z$ to state $z'$. The associated generator acts as
\[
\cL_t f(z) = \sum_{z' \neq z} R_t(z,z' \mid x_1)\bigl(f(z') - f(z)\bigr),
\]
or analogously as an integral when the state space is continuous.

\subsection{Auxiliary Generator Matching Proofs}\label{app:aux-gen-proofs}
\begin{revblock}
The theorems in this section hold under the following assumptions.

\begin{assumption}[Joint process]\label{ass:joint-process}
    $(X_t,S_t)_{t\in[0,1]}$ is a Markov process on $\bX\times\bS$ with marginals $p_t(x,s)$ and generator $\cL_t$. For every test function $g$ (Assumption~\ref{ass:test-functions}), the map $t\mapsto\int g\,dp_t$ is differentiable and satisfies the weak Fokker-Planck equation \eqref{eq:fokker-planck-weak}. This holds in the setting of Generator Matching, i.e.\ for Feller processes with finitely many expected jumps on $[0,1]$ and with $t\mapsto\cL_tg$ and $t\mapsto\int g\,dp_t$ continuous \citep[Appendix~A.2, Assumptions~1--4]{holderrieth2024generator}.
\end{assumption}

\begin{assumption}[Test functions]\label{ass:test-functions}
    We call $g$ a \emph{test function} of $\cL_t$ if $g$ lies in the domain of $\cL_t$ and $\cL_tg\in L^1(p_t)$. The domain of $\cK_t$ is the class $\cT$ of functions $f:\bX\to\bR$ such that $\varphi_f$ is a test function of $\cL_t$. We assume that $\cT$ determines probability measures on $\bX$: $\mu_1=\mu_2$ if and only if $\int f\,d\mu_1=\int f\,d\mu_2$ for all $f\in\cT$ \citep[Appendix~A.2, Assumption~3]{holderrieth2024generator}.
\end{assumption}

\begin{assumption}[Integrable targets]\label{ass:integrable-targets}
    The conditional target is integrable under the joint law of the data and the process at time $t$:
    \[
        \bE_{(x_1,s_1)\sim p_{\data}}\,\bE_{(x_t,s_t)\sim p_{t|1}(\cdot\mid x_1,s_1)}\big[\|F_{t,x_1,s_1}(x_t,s_t)\|\big]<\infty .
    \]
    By Fubini's theorem, the conditional expectation $F_t(x)=\bE_{x_1,s_1,s_t}[F_{t,x_1,s_1}(x,s_t)\mid X_t=x]$ then exists for $p_t$-almost every $x$. This assumption is only needed for the linear parametrization.
\end{assumption}
\end{revblock}

We start by giving the proof of theorem \ref{thm:joint-generators}.

\begin{proof}
    We first show that $\cK_t$ generates $p_t(x)$ by showing it satisfies the Fokker Planck equation:
    \begin{align*}
        \partial_t\int_{\bX} f(x) p_t(x) &=  \partial_t \int_{\bX\times\bS} \varphi_f(x,s) p_t(x,s)\\
        &= \int_{\bX\times \bS} \cL_t \varphi_f(x,s) p_t(x,s) \\
        &= \int_{\bX} p_t(x) \int_{\bS} \cL_t \varphi_f (x,s) p_t(s_t|X_t=x) \\
        &= \int_{\bX} p_t(x) \cK_t f(x) 
    \end{align*}
    \rev{In the third line we have used that $\cL_t\varphi_f\in L^1(p_t)$ by Assumption~\ref{ass:test-functions}, so by Fubini's theorem $\cL_t\varphi_f(x,S_t)$ is integrable given $X_t=x$ for $p_t$-almost every $x$; hence $\cK_tf$ is well defined $p_t$-almost everywhere and the integrals can be split.}
    This proves the first part of the theorem. The factorization principle \rev{follows from the representation above, equation \eqref{eq:joint-generator-factorization} and the tower property:}
    \begin{revblock}
    \begin{align*}
        \cK_t f(x) &= \bE_{s_t}[ \cL_t \varphi_f(x,s_t) \mid X_t=x] \\
        &= \bE_{s_t}\Big[ \bE_{x_1, s_1} \big[ \cL_{t,x_1,s_1} \varphi_f(x,s_t) \mid X_t=x, S_t = s_t\big] \,\Big|\, X_t = x\Big] \\
        &= \bE_{x_1,s_1,s_t}\big[ \cL_{t,x_1,s_1} \varphi_f(x,s_t) \mid X_t=x\big]
        = \bE_{x_1,s_1,s_t}\big[ \cK_{t,x_1,s_1,s_t} f(x) \mid X_t=x\big].
    \end{align*}
    For the linear parametrization, the conditional parametrization in \eqref{eq:joint-generator-linear} and the assumption $K\varphi_f(x,s)=K_xf(x)$ give $\cK_{t,x_1,s_1,s_t}f(x)=\langle K_xf(x),F_{t,x_1,s_1}(x,s_t)\rangle$. The first argument does not depend on $(x_1,s_1,s_t)$, so by the factorization above, linearity of the pairing and Assumption~\ref{ass:integrable-targets},
    \begin{align*}
        \cK_t f(x) &= \bE_{x_1,s_1,s_t}\big[\langle K_xf(x),F_{t,x_1,s_1}(x,s_t)\rangle \mid X_t=x\big] \\
        &= \Big\langle K_xf(x),\,\bE_{x_1,s_1,s_t}\big[F_{t,x_1,s_1}(x,s_t)\mid X_t=x\big]\Big\rangle \\
        &= \langle K_xf(x),F_t(x)\rangle .
    \end{align*}
    Notice that no extra assumption is needed since $F_t$ is defined this way, and the quantity exists by the integrability of $F_{t,x_1,s_1}$.
    \end{revblock}
\end{proof}

\begin{revblock}
\begin{remark}
    The statement that $\cK_t$ \emph{generates} $p_t(x)$ means that $p_t(x)$ solves the weak Fokker-Planck equation of $\cK_t$ for all $f\in\cT$. Simulating $\cK_t$ recovers $p_t(x)$ if $\cK_t$ is the generator of a Markov process for which the weak Fokker-Planck equation identifies the path \citep[Appendix~A.2, Assumption~5]{holderrieth2024generator}; in that case, this process started from $p_0$ has marginals $p_t(x)$. 
\end{remark}
\end{revblock}

\begin{remark}
    The theorem above gives us a direct construction of how to compute a marginalized process, as well as sufficient conditions under which such process can be linearized, so that it can be learnable under the framework of generator matching.
\end{remark}
\begin{remark}
    A similar theorem in \cite{billera2025branching} was used to construct their process. We must emphasize that the theorems are different. Intuitively, their approach allows for conditioning on a \emph{discrete} process, on the other hand, instead of conditioning on a process, we consider a joint process over an \emph{arbitrary} state space that we marginalize over. Mathematically, \citet{billera2025branching} assume that (in our notation) for each $s \in \bS$ there exists a generator $\cL_t^s$ such that:
    \[\partial_t \langle f(x), p_t(x,s) \rangle = \langle \cL_t^s f(x), p_t(x,s) \rangle\]
    That is, it assumes that there exists a generator for the process under a fixed $s$. On the other hand, we assume that we have a generator over a joint space $\bX \times \bS$ and directly construct a process that is learnable through generator matching, which simplifies the design of the auxiliary process.
\end{remark}
\begin{remark}
    We would like to comment on the assumption of $K \varphi_f (x,s) = K_x f(x)$. This is not restrictive as it corresponds to operators such that $K$ affects coordinates locally, while the coupling between variables is captured by the function $F$. This class admits coupled dynamics—including flows, diffusions, and joint jumps whose coefficients depend on latent variables—but excludes operators that mix coordinates. For example generators of the form $\cL_t f(x,s) = \partial_{x,s} f(x,s)$ do not satisfy this assumption. Under this assumption, marginalization of the $s$ variable preserves the operator structure.
\end{remark}

We now present the proof that Bregman divergences are a suitable training objective as in Theorem \ref{thm:loss-function}.\

\begin{proof}
\rev{By Theorem~\ref{thm:joint-generators}, the marginal generator is parametrized as $\cK_t f(x)=\langle K_xf(x),F_t(x)\rangle$, where the target is defined by}
\[
    F_t(x)
    =
    \mathbb{E}_{x_1,s_1,s_t}
    \big[
        F_{t,x_1,s_1}(x,s_t)
        \,\big|\, X_t = x
    \big].
\]
Let $D$ be the Bregman divergence generated by a convex function $\phi$ \rev{and consider parameters $\theta$ for which the loss $L_{\mathrm{cgm-aux}}(\theta)$ is finite},
\[
    D(a,b)
    =
    \phi(a)
    -
    \phi(b)
    -
    \langle a-b, \nabla \phi(b) \rangle .
\]
Then
\begin{align*}
    L_{\mathrm{gm}}
    &=
    \mathbb{E}_t
    \mathbb{E}_{x_t}
    \big[
        D(F_t(x_t), F_t^\theta(x_t))
    \big] \\
    &=
    \mathbb{E}_t
    \mathbb{E}_{x_t}
    \Big[
        \phi(F_t(x_t))
        -
        \phi(F_t^\theta(x_t))
        -
        \langle
            F_t(x_t) - F_t^\theta(x_t),
            \nabla \phi(F_t^\theta(x_t))
        \rangle
    \Big].
\end{align*}

Substituting the conditional expectation representation of $F_t(x_t)$ and
applying the tower property yields
\begin{align*}
    L_{\mathrm{gm}}
    &=
    \mathbb{E}_t
    \mathbb{E}_{x_1,s_1}
    \mathbb{E}_{x_t,s_t \mid x_1,s_1}
    \Big[
        \phi(F_{t,x_1,s_1}(x_t,s_t))
        -
        \phi(F_t^\theta(x_t)) \\
    &\qquad\qquad
        -
        \langle
            F_{t,x_1,s_1}(x_t,s_t) - F_t^\theta(x_t),
            \nabla \phi(F_t^\theta(x_t))
        \rangle
    \Big]
    + C,
\end{align*}
where
\[
    C
    =
    \mathbb{E}_t
    \mathbb{E}_{x_t}
    \big[
        \phi(F_t(x_t))
        -
        \mathbb{E}_{x_1,s_1,s_t \mid X_t=x_t}
        \phi(F_{t,x_1,s_1}(x_t,s_t))
    \big]
\]
is independent of $\theta$.

Therefore,
\[
    L_{\mathrm{gm}}
    =
    \mathbb{E}_t
    \mathbb{E}_{x_1,s_1}
    \mathbb{E}_{x_t,s_t \mid x_1,s_1}
    \big[
        D(F_{t,x_1,s_1}(x_t,s_t), F_t^\theta(x_t))
    \big]
    + C,
\]
with $C$ independent of $\theta$ and the loss functions have the same gradients.
\end{proof}

\begin{revblock}
\begin{remark}[Reweighting]\label{rem:reweighting}
    If the target has several components, $F=(F^1,\dots,F^m)$, and $D^1,\dots,D^m$ are Bregman divergences generated by $\phi^1,\dots,\phi^m$, then for any weights $w^j_t>0$ the sum $D_t=\sum_j w^j_tD^j$ is the Bregman divergence generated by the convex function $\phi_t(a)=\sum_j w^j_t\phi^j(a^j)$. By Theorem~\ref{thm:loss-function}, any positive time-dependent reweighting of the terms of the conditional loss therefore has the same gradient as the correspondingly reweighted marginal loss.
\end{remark}
\end{revblock}

\subsection{Proofs for our Process}\label{app:process-proofs}
\begin{revblock}
In the interpolant construction of Section~\ref{sec:practical-process} the coordinates do not interact: each one waits, is inserted, and is then denoised, independently of the others. We therefore first describe a single coordinate and then combine the coordinates to prove Theorem~\ref{thm:conditional-prob-path-factorization}. Throughout, $\gamma$ is continuously differentiable on $\{t:\gamma_t<1\}$ and $\lambda_t:=\dot\gamma_t/(1-\gamma_t)$ denotes its hazard rate.

\begin{lemma}[One coordinate]\label{lem:single-coordinate}
Fix $v\in\tilde\bV$. Let $\tau$ be a random time with $\bP(\tau\le t)=\gamma_t$ and, independently of $\tau$, let $(Y_u)_{u\in[0,1]}$ be a Markov process with right-continuous paths, generator $\cA_{u,v}$ and $Y_0\sim q(\cdot\mid v)$, so that $Y_u\sim p_u(\cdot\mid v)$. Define the process
\[
    Z_t:=\begin{cases}\varnothing, & t<\tau,\\ \bigl(Y_{u(t;\tau)},\,\tau\bigr), & t\ge\tau,\end{cases}
\]
with values in $\{\varnothing\}\cup(\tilde\bV\times[0,1])$, where $\varnothing$ stands for ``not yet inserted''. Then:
\begin{enumerate}
    \item[(a)] $Z_t=\varnothing$ with probability $1-\gamma_t$, and otherwise $Z_t=(y,\tau)$ has density $\dot\gamma_\tau\,p_{u(t;\tau)}(y\mid v)$ on $\tau\in[0,t]$.
    \item[(b)] $Z$ is a Markov process and, whenever $\gamma_t<1$, its generator $\cG_{t,v}$ is given by
    \[
        \cG_{t,v}\,g(\varnothing)=\lambda_t\int\bigl[g(y,t)-g(\varnothing)\bigr]\,q(dy\mid v),
        \qquad
        \cG_{t,v}\,g(y,\tau)=\frac{1}{1-\tau}\,\bigl(\cA_{u(t;\tau),v}\,g(\cdot,\tau)\bigr)(y).
    \]
\end{enumerate}
\end{lemma}

\begin{proof}
(a) We have $\bP(\tau>t)=1-\gamma_t$ and $\tau$ has density $\dot\gamma_\tau$. Given $\tau\le t$, the value $Y_{u(t;\tau)}$ has law $p_{u(t;\tau)}(\cdot\mid v)$, since $Y$ is independent of $\tau$.

(b) \emph{Markov property.} If $Z_t=\varnothing$, the past of $Z$ only reveals that $\tau>t$, which is the same information as $Z_t=\varnothing$. If $Z_t=(y,\tau)$, the insertion time is part of the state and, since $Y$ is Markov and independent of $\tau$, the future of the value depends on its past only through $y$. In both cases the future of $Z$ depends on the past through $Z_t$ alone.

\emph{Generator.} If $Z_t=\varnothing$, the coordinate is inserted in $(t,t+h]$ with probability
\[
    \bP(\tau\le t+h\mid\tau>t)=\frac{\gamma_{t+h}-\gamma_t}{1-\gamma_t}=\lambda_t h+o(h),
\]
in which case $Z_{t+h}=(Y_{u(t+h;\tau)},\tau)$ converges to $(Y_0,t)$ as $h\to0$ by right-continuity, with $Y_0\sim q(\cdot\mid v)$. This gives the first formula. If $Z_t=(y,\tau)$, the value is the process $Y$ run on the clock $u(t;\tau)$, so its generator is that of $Y$ multiplied by the speed of the clock, $\partial_t u(t;\tau)=1/(1-\tau)$. This gives the second formula.
\end{proof}

\begin{proof}[Proof of Theorem~\ref{thm:conditional-prob-path-factorization}]
Fix $x_1\in\tilde\bV^L$, so that $s_1=[L]$. Let $Z^1,\dots,Z^L$ be independent, where $Z^i$ is the process of Lemma~\ref{lem:single-coordinate} with $v=x_1^{(i)}$ and insertion time $\tau_i$. Steps (i)--(ii) of Section~\ref{sec:practical-process} define exactly
\[
    S_t=\{i:Z^i_t\neq\varnothing\},\qquad X_t=(Z^i_t)_{i\in S_t},
\]
that is, $X_t$ lists the inserted coordinates in order and $S_t$ records their indices. The pair $(X_t,S_t)$ and the tuple $(Z^1_t,\dots,Z^L_t)$ determine each other, so they are two ways of writing the same state.

Independent Markov processes are jointly Markov. Their joint law is the product of the individual laws and, by the product rule, their generator is the sum of the individual generators, each acting on its own coordinate. Hence $(X_t,S_t)$ is Markov, and it remains to write this law and this generator in terms of $(x,S)$.

By Lemma~\ref{lem:single-coordinate}(a), every $j\notin S_t$ contributes a factor $1-\gamma_t$ and every $i\in S_t$ a factor $\dot\gamma_{\tau_i}\,p_{u(t;\tau_i)}(x_t^{(i)}\mid x_1^{(i)})$, so
\[
    p_t(x_t,S_t\mid x_1,S_1)=\underbrace{(1-\gamma_t)^{L-|S_t|}\prod_{i\in S_t}\dot\gamma_{\tau_i}}_{p_t(S_t,\tau\mid S_1)}\;\prod_{i\in S_t}p_{u(t;\tau_i)}\bigl(x_t^{(i)}\mid x_1^{(i)}\bigr),
\]
which is \eqref{eq:conditional-prob-path-factorization}.

\emph{Generator.} By Lemma~\ref{lem:single-coordinate}(b), every $i\in S$ contributes the base generator of its value at speed $1/(1-\tau_i)$, and the sum of these terms is $\Ldenoise_{t,x_1}$ in \eqref{eq:denoise-generator}. Every $j\notin S$ contributes a jump of $Z^j$ from $\varnothing$ to $(v,t)$ at rate $\lambda_t$, with $v\sim q(\cdot\mid x_1^{(j)})$. In terms of $(x,S)$, this single jump inserts the entry $(v,t)$ into $x$ at the position $k(j)$ with $s^{(k(j)-1)}<j<s^{(k(j))}$ and adds $j$ to $S$. The sum of these terms is therefore
\[
    \lambda_t\sum_{j\notin S}\int\Bigl[f\bigl(\ins_{k(j)}(x,v,t),\,S\cup\{j\}\bigr)-f(x,S)\Bigr]\,q\bigl(dv\mid x_1^{(j)}\bigr)=\Linsert_{t,x_1,s_1}f(x,S),
\]
the jump generator with kernel \eqref{eq:insertion-kernel}.
\end{proof}
\end{revblock}

Finally, we prove Lemma~\ref{lemma:loss-for-our-case}:

\begin{proof}
Recall that the conditional generator is $\cL_{t,x_1,s_1}=\Ldenoise_{t,x_1}+\Linsert_{t,x_1,s_1}$, and that each component admits a linear parametrization in the sense of \eqref{eq:joint-generator-linear}. Applying Theorem~\ref{thm:loss-function} yields the objective
\[
\mathbb{E}_{t,(x_1,s_1),(x_t,s_t)}\Big[D\!\left(F_{t,x_1,s_1}(x_t,s_t), F_t^\theta(x_t)\right)\Big],
\]
and since the generator decomposes additively, the feature representation, and hence the loss, separates into a denoising and an insertion contribution. The denoising term is $D\big(F^{\text{denoise}}_{t,x_1,s_1}(x_t,s_t),F^{\theta,\text{denoise}}_t(x_t)\big)$.

The insertion component is a jump process. Write $\lambda_t:=\dot\gamma_t/(1-\gamma_t)$ and $m_t[j]:=s_t[j]-s_t[j-1]-1$ for the number of missing coordinates in gap $j$. On $\bX$, the state reached by inserting a missing coordinate $i$ of gap $j$ with value $v$ is $\ins_j(x_t,v,t)$, which does not depend on $i$. Summing the kernel \eqref{eq:insertion-kernel} over these coordinates, the conditional jump measure out of gap $j$ is
\[
\mu_{t,j}(dv)=\lambda_t\sum_{s_t[j-1]<i<s_t[j]} q\bigl(dv\mid x_1^{(i)}\bigr),
\]
with total mass $\lambda_t\, m_t[j]$. We parametrize it as $\mu^\theta_{t,j}(dv)=\lambda_t\,F^{\theta,\text{insert}}_t(x_t,j)\,q^\theta_t(dv\mid x_t,j)$ with $q^\theta_t$ normalized, and take as Bregman divergence the generalized KL between jump measures, $D(\mu,\mu^\theta)=\int\big(d\mu^\theta-\log\tfrac{d\mu^\theta}{d\nu}\,d\mu\big)$ up to terms independent of $\theta$, for a reference measure $\nu$. Substituting,
\[
D\big(\mu_{t,j},\mu^\theta_{t,j}\big)
=\lambda_t\,\phi\!\left(m_t[j],F^{\theta,\text{insert}}_t(x_t,j)\right)
+\lambda_t\sum_{s_t[j-1]<i<s_t[j]}\bE_{v\sim q(\cdot\mid x_1^{(i)})}\!\left[-\log q^\theta_t(v\mid x_t,j)\right]+C,
\]
where $C$ does not depend on $\theta$. The first term is the insertion loss and the second the initialization loss; when $q$ is a prior, $q^\theta_t=\pi$ is fixed and the second term is constant. \rev{Summing over gaps, Theorem~\ref{thm:loss-function} gives the stated objective with the insertion and initialization terms multiplied by $\lambda_t>0$. Dividing these terms by $\lambda_t$ is a positive time reweighting, so by Remark~\ref{rem:reweighting} the stated objective is again a Bregman divergence loss and Theorem~\ref{thm:loss-function} applies to it directly.}
\end{proof}

\section{Design choices for our method}
\subsection{Insertion Methods} \label{app:insertion-methods}

As discussed in Section~\ref{sec:practical-process}, the insertion kernel only adds positions to $S_t$ (never removes them) and can be specified independently of the denoising dynamics. All variants below share the same global schedule $\gamma_t$, which controls the \emph{rate} at which insertions happen; they differ in \emph{which} positions receive the next insertion. The list below is illustrative, not exhaustive: any rate matrix that monotonically grows $S$ to the target size yields a valid variable-length generator, and the choice can be tailored to the structure of the data.

\textit{Random insertion.} The default kernel from Equation~\eqref{eq:full-forward-process} of the main body, repeated here for reference:
\begin{align*}
R_t\bigl((x,S), (\ins(x,i), S\cup\{i\}) \mid x_1\bigr)
= \frac{\dot \gamma_t}{1-\gamma_t}\,\indicator\{s^{(i-1)} < i < s^{(i)}\},
\end{align*}
with boundary conventions $s^{(0)} = 0$ and $s^{(|S|+1)} = \len(x_1)+1$. The main appeal is order-agnosticism: the model can fill in the structure in whatever order best matches the data.

\textit{Auto-regressive insertion.} Alternatively, one can restrict insertions to the right end of the current sequence, recovering an auto-regressive growth pattern:
\begin{align*}
R_t\bigl((x,S), (\ins(x,\len(x)+1), S\cup\{\len(x)+1\}) \mid x_1\bigr)
= \frac{\dot \gamma_t}{1-\gamma_t}\,(\len(x_1) - \len(x)).
\end{align*}
The $(\len(x_1) - \len(x))$ factor scales the rate by the remaining length, so the expected number of insertions tracks the global schedule $\gamma_t$ regardless of how far the sequence has grown. This kernel is natural when the underlying data has an inherent left-to-right ordering.

\textit{Block insertion.} Both kernels above can be extended to insert several mask tokens at once, which is useful when the data evolves in fixed-size groups --- for example, $3$D coordinates that come in blocks of three, or block-wise schemes like Block Diffusion \cite{arriola2025block}.

\subsection{Deletions via Junk Augmentation} \label{app:deletions}

The recipe of Section~\ref{sec:practical-process} extends to processes that can also \emph{delete} coordinates, through the standard procedure that we outline here. The idea is to augment each data point with junk coordinates that are deleted before $t=1$, so that the model learns which coordinates do not belong. We spell out the construction and show that Theorem~\ref{thm:conditional-prob-path-factorization} and Lemma~\ref{lemma:loss-for-our-case} carry over unchanged.

\paragraph{Augmentation.} Given $x_1\in\tilde\bV^{L}$, sample $(\tilde x_1,\ell)\sim q_{\rm aug}(\cdot\mid x_1)$, where $\tilde x_1\in\tilde\bV^{L+J}$ is obtained by inserting $J\ge 0$ junk values into $x_1$ and $\ell\in\{\text{real},\text{junk}\}^{L+J}$ labels each slot; write $\cJ:=\{i:\ell_i=\text{junk}\}$. Removing the slots in $\cJ$ from $\tilde x_1$ recovers $x_1$. The augmentation distribution is a design choice; a natural default places a Poisson number of junk slots uniformly among the gaps of $x_1$ and draws their values from the marginal distribution of coordinate values, so that junk cannot be recognized from its value alone. The case $J=0$ recovers the insertion-only process of the main text.

\paragraph{Lifetimes.} Fix increasing schedules $\gamma_t$ and $\beta_t$ with $\gamma_0=0$ and $\gamma_1=\beta_1=1$. Each slot $i\in[L+J]$ receives an insertion time $\tau_i$ with $\bP(\tau_i>t)=1-\gamma_t$, as in the main text, and each junk slot additionally receives a deletion time $\delta_i>\tau_i$ with $\bP(\delta_i>t\mid\tau_i)=(1-\beta_t)/(1-\beta_{\tau_i})$; real slots have $\delta_i=\infty$. Slot $i$ is alive during $[\tau_i,\delta_i)$. The auxiliary variable records the alive slots and the slots already deleted,
\begin{align*}
    s_t=(S_t,\Delta_t),\qquad S_t=\{i:\tau_i\le t<\delta_i\}\ \text{(ordered)},\qquad \Delta_t=\{i\in\cJ:\delta_i\le t\},
\end{align*}
and slots in neither set have not been inserted yet. Since $\gamma_1=\beta_1=1$, at $t=1$ every real slot is alive and every junk slot has been deleted, so the process ends at $x_1$. While alive, every slot follows the base interpolant on its local clock exactly as in the main text, $x_t^{(i)}\sim p_{u(t;\tau_i)}(\cdot\mid\tilde x_1^{(i)})$; junk slots are denoised toward their junk value, so in isolation they are statistically indistinguishable from real ones and can only be identified through their context.

\paragraph{Generator.} The conditional generator gains a deletion term,
\begin{align}\label{eq:full-forward-process-deletions}
    \cL_{t,\tilde x_1,\ell}=\Linsert_{t,\ell}+\Ldenoise_{t,\tilde x_1}+\Ldelete_{t,\ell}.
\end{align}
Write $z=(x,S,\Delta)$, let $z^{+}_{k,i}(v):=(\ins_k(x,v,t),\,S\cup\{i\},\,\Delta)$ denote the state after inserting slot $i$ at position $k$ with value $v$ and clock $t$, and let $z^{-}_{k}:=(\delete_k(x),\,S\setminus\{s^{(k)}\},\,\Delta\cup\{s^{(k)}\})$ denote the state after removing the $k$-th coordinate together with its clock. Insertion uses the kernel \eqref{eq:insertion-kernel} restricted to slots that have not been deleted, and deletion fires only on junk:
\begin{align}
    R^{\ins}_t\bigl(z\to z^{+}_{k,i}(v)\bigr)&=\frac{\dot\gamma_t}{1-\gamma_t}\,\indicator\{s^{(k-1)}<i<s^{(k)},\ i\notin\Delta\}\,q\bigl(v\mid\tilde x_1^{(i)}\bigr),\\
    R^{\delete}_t\bigl(z\to z^{-}_{k}\bigr)&=\frac{\dot\beta_t}{1-\beta_t}\,\indicator\{s^{(k)}\in\cJ\}. \label{eq:deletion-kernel}
\end{align}
A junk slot inserted at time $\tau$ survives until time $t$ with probability $(1-\beta_t)/(1-\beta_\tau)$, which vanishes at $t=1$, and a deleted slot is never re-inserted. The denoising generator is \eqref{eq:denoise-generator} with targets $\tilde x_1$. Real and junk slots are inserted through the same kernel, so the insertion mechanism cannot distinguish them.

\subsection{Sampler Pseudocode} \label{app:sampler-pseudocode}
We provide pseudocode for the two higher-order sampler variants from Section~\ref{sec:higher-order-samplers}.
\noindent
\begin{minipage}[t]{0.48\linewidth}
\begin{codeblock}
\begin{minted}[fontsize=\scriptsize,highlightcolor=myhighlight,baselinestretch=0.96]{python}
def sample(x, N):  # x: initial condition, N: num steps
    h = 1 / N
    t1 = 0
    t2 = 0
    x, t2 = flow_A(x, t1, h / 2), h/2
    for n in range(N):
        x = flow_B(x, t2, h)
        t1 += h
        dt = h if n != N-1 else h/2
        x = flow_A(x, t1, dt)
        t2 += dt
    return x
\end{minted}
\end{codeblock}
\captionof{listing}{Staggered operator-splitting sampler (NFEs = 2).}
\label{code:staggered}
\end{minipage}
\hfill
\begin{minipage}[t]{0.48\linewidth}
\begin{codeblock}
\begin{minted}[fontsize=\scriptsize,highlightcolor=myhighlight,baselinestretch=0.96]{python}
def sample(x, N):  # x: initial condition, N: num steps
    h = 1 / N
    t1 = 0
    t2 = 0
    for n in range(N):
        x = flow_A(x, t1, h / 2)
        t2 += h / 2
        x = flow_B(x, t2, h)
        t1 += h
        x = flow_A(x, t1, h / 2)
        t2 += h / 2
    return x
\end{minted}
\end{codeblock}

\captionof{listing}{Direct operator-splitting sampler (NFEs = 3).}
\label{code:split}
\end{minipage}

\begin{revblock}
\subsection{Empirical Order of the Samplers} \label{app:sampler-order}
Figure~\ref{fig:error-vs-h} reports, for each sampler, the error minus the best error reached by any sampler, against the step size $h=1/N$. On QM9, the fraction of invalid and of disconnected molecules decreases at a slope of about $1$ for all three samplers, with nearly identical constants. On RICO, all samplers, including Euler, decrease at a slope of about $2$ in FID; this is expected, since FID is a squared distance between feature statistics, so that a bias of order $h$ in the samples appears at order $h^2$. The splitting samplers therefore converge at the same empirical order as Euler.

Although Strang splitting is second order only if every split flow is applied exactly (Section~\ref{sec:higher-order-samplers}). In our samplers, this is not possible, due to the simulation of CTMCs, on the other hand for Euclidean diffusion we are able to integrate them exactly which we do in practice. Nevertheless, we observe that the splitting lowers the constant: on RICO, the direct split reduces the excess FID by a factor of $3$ to $4$ at $5$ to $10$ steps (Figure~\ref{fig:rico-samplers}). 

\begin{figure}[h]
    \centering
    \begin{revblock}
    \includegraphics[width=\linewidth]{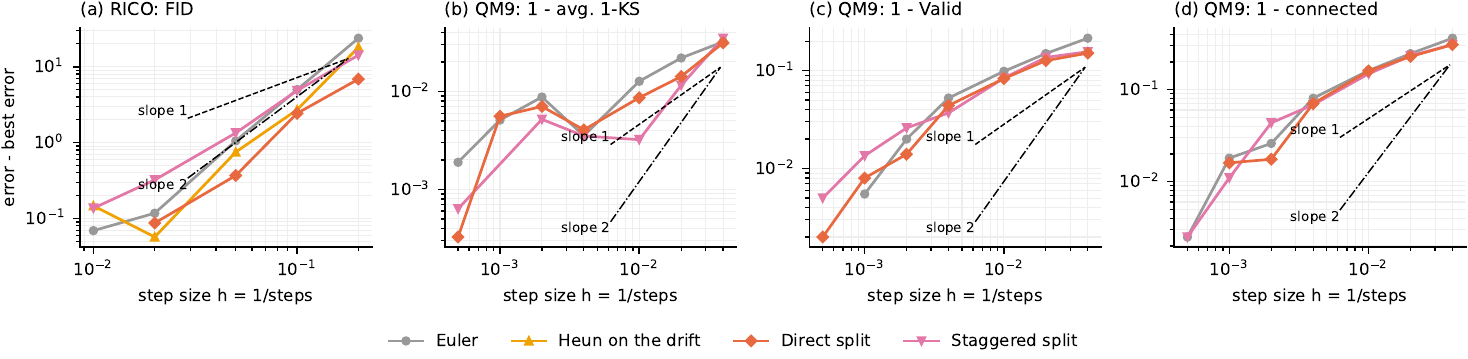}
    \caption{\textbf{Error versus step size}, minus the best error reached by any sampler (log-log). (a) RICO, FID . (b--d) QM9: one minus the average $1-$KS, the fraction of valid molecules and the fraction of connected molecules ($2$K molecules from one sampling seed per point). Dashed and dash-dotted lines have slopes $1$ and $2$.}
    \label{fig:error-vs-h}
    \end{revblock}
\end{figure}
\end{revblock}

\section{Experimental Details} \label{app:hyperparameters}

In this section we present experimental details for the experiments. Every model was trained and evaluated on a single NVIDIA A100 (80\,GB) GPU. Hyperparameters for RICO and PubLayNet are in Table~\ref{tab:hyperparameters-layout}.

\begin{table}[h]
    \centering
    \caption{Hyperparameters for layout generation on RICO and PubLayNet. The backbone matches the size of the Transformer used by \citet{guerreiro2024layoutflow}.}
    \label{tab:hyperparameters-layout}
    \setlength{\tabcolsep}{6pt}
    \renewcommand{\arraystretch}{1.2}
    \small
    \resizebox{\linewidth}{!}{%
    \begin{tabular}{lcc}
        \toprule
        \textbf{Hyperparameter} & \textbf{RICO} & \textbf{PubLayNet} \\
        \midrule
        Architecture        & \multicolumn{2}{c}{Transformer over the element set, no positional encoding} \\
        Layers & 4 & 4\\
        Width & 512  & 512 \\
        Heads & 8  & 8\\
        Parameters          & 15.7M & 15.7M \\
        Insertion           & from the prior, own clock & \rev{from the prior, own clock} \\
        \rev{Training tasks} & \multicolumn{2}{c}{\rev{4 conditional tasks + unconditional, equal parts of each batch}} \\
        Reveal horizon $t_{\max}$ & 0.5 & 0.5 \\
        Loss weights (category / insertion / $\ell_1$) & 0.25 / 0.1 / 0.2 & 0.25 / 0.1 / 0.2 \\
        Training            & 2{,}000 epochs (124K steps) & 1{,}000 epochs (609K steps) \\
        Batch size          & 512 & 512 \\
        Optimizer           & AdamW & AdamW \\
        Learning rate       & \multicolumn{2}{c}{$5\times10^{-4}$, reduced $10\times$ on a validation-FID plateau} \\
        EMA                 & none & none \\
        Integration steps (eval) & 100 (Euler) & 100 (Euler) \\
        \bottomrule
    \end{tabular}}
\end{table}

\subsection{Layout Generation: Additional Details} \label{app:layout-details}

\paragraph{Data representation.}
Each element is a pair of a category in $[C]$ and a bounding box in $[0,1]^4$, parametrized by center and size and normalized by the canvas dimensions. Layouts with more than $20$ elements are discarded, following \citet{zhang_layoutdiffusion,jiang2023layoutformer++}. There are $C=25$ categories on RICO and $5$ on PubLayNet. The layout is treated as a set: the backbone has no positional encoding and elements that do not exist yet are hidden by a padding mask, so the insertion kernel adds elements without an ordering. \rev{On both datasets the model inserts an element with a rate per existing element and draws its box from the Gaussian prior on its own clock. A single model is trained to complete all tasks: each training batch is split in five equal parts, one per conditional task (\tasktype, \tasktypesize, completion and refinement) and one for unconditional generation, and the unconditional and conditional results are all obtained from this model.}

\paragraph{Evaluation.}
We follow the same evaluation protocol as in the official respository of \citet{zhang_layoutdiffusion}. Including the task types, architecture and data splits.

\begin{table}[h]
    \centering
    \caption{\textbf{Layout generation on RICO and PubLayNet, all tasks and methods.} Baseline numbers are from \citet{guerreiro2024layoutflow};}
    \label{tab:layout-results-full}
    \resizebox{\linewidth}{!}{
    \begin{tabular}{llrrrrrrrrr}
        \toprule
        & & \multicolumn{4}{c}{RICO} && \multicolumn{4}{c}{PubLayNet}\\
        \cmidrule{3-6}\cmidrule{8-11}
        Task & Model & \abbfid & \abbalignment & \abboverlap & \abbmiou & $\quad$ & \abbfid & \abbalignment & \abboverlap & \abbmiou \\
        \midrule
        \multirow{7}{*}{\taskugen} & LayoutTransformer & 24.32 & 0.037 & 0.542 & 0.587 && 30.05 & 0.067 & 0.005 & 0.359\\
        & LayoutFormer++    & 20.20 & 0.051 & 0.546 & 0.634 && 47.08 & 0.228 & 0.001 & 0.401\\
        \cmidrule{2-11}
        & LayoutDM$^{*\ddagger}$   & 4.43  & 0.143 & 0.584 & 0.582 && 36.85 & 0.180 & 0.132 & 0.382\\
        & DLT$^{*\dagger}$         & 13.02 & 0.271 & 0.571 & 0.566 && 12.70 & 0.117 & 0.036 & 0.431\\
        & LayoutDiffusion$^{\ddagger}$ & 2.49  & 0.069 & 0.502 & 0.620 && 8.63  & 0.065 & 0.003 & 0.417\\
        & LayoutFlow$^{\dagger}$   & 2.37  & 0.150 & 0.498 & 0.570 && 8.87  & 0.057 & 0.009 & 0.424\\
        \rowcolor{myhighlight} & VL-DE (ours)       & 2.32  & 0.253 & 0.547 & 0.578 && \rev{8.45}  & \rev{0.130} & \rev{0.014} & \rev{0.421}\\
        \midrule
        \multirow{7}{*}{\tasktype} & NDN-none & 13.76 & 0.560 & 0.550 & 0.350 && 35.67 & 0.350 & 0.170 & 0.310\\
        & LayoutFormer++    & 2.48  & 0.124 & 0.537 & 0.377 && 10.15 & 0.025 & 0.009 & 0.333\\
        \cmidrule{2-11}
        & LayoutDM$^*$       & 2.39  & 0.222 & 0.598 & 0.341 && 39.12 & 0.267 & 0.139 & 0.348\\
        & DLT$^*$            & 6.64  & 0.303 & 0.616 & 0.326 && 7.09  & 0.097 & 0.040 & 0.349\\
        & LayoutDiffusion    & 1.56  & 0.124 & 0.491 & 0.345 && 3.73  & 0.029 & 0.005 & 0.343\\
        & LayoutFlow         & 1.48  & 0.176 & 0.517 & 0.322 && 3.66  & 0.037 & 0.011 & 0.350\\
        \rowcolor{myhighlight} & VL-DE (ours)       & 1.36  & 0.177 & 0.523 & 0.339 && \rev{3.64} & \rev{0.058} & \rev{0.015} & \rev{0.351}\\
        \midrule
        \multirow{4}{*}{\tasktypesize} & LayoutDM$^*$ & 1.76 & 0.175 & 0.606 & 0.424 && 29.91 & 0.246 & 0.160 & 0.436\\
        & DLT$^*$            & 6.27  & 0.332 & 0.609 & 0.424 && 5.35  & 0.130 & 0.053 & 0.426\\
        & LayoutFlow         & 1.03  & 0.283 & 0.523 & 0.470 && 1.26  & 0.041 & 0.031 & 0.454\\
        \rowcolor{myhighlight} & VL-DE (ours)       & 1.11  & 0.193 & 0.525 & 0.453 && \rev{2.28} & \rev{0.067} & \rev{0.041} & \rev{0.425}\\
        \midrule
        & \textbf{Validation data} & 2.10 & 0.093 & 0.466 & 0.658 && 8.10 & 0.022 & 0.003 & 0.434\\
        \bottomrule
    \end{tabular}
    }
\end{table}

\begin{table}[h]
    \centering
    \caption{\textbf{Completion and refinement on RICO and PubLayNet.} Baseline numbers are from \citet{guerreiro2024layoutflow}; models marked with $*$ were retrained there on the common split. In completion, LayoutFlow and LayoutDM are given the number of missing elements, whereas VL-DE generates it.}
    \label{tab:layout-completion-refinement}
    \resizebox{\linewidth}{!}{
    \begin{tabular}{llrrrrrrrrr}
        \toprule
        & & \multicolumn{4}{c}{RICO} && \multicolumn{4}{c}{PubLayNet}\\
        \cmidrule{3-6}\cmidrule{8-11}
        Task & Model & \abbfid & \abbalignment & \abboverlap & \abbmiou & $\quad$ & \abbfid & \abbalignment & \abboverlap & \abbmiou \\
        \midrule
        \multirow{3}{*}{\taskcompletion} & LayoutDM$^*$ & 6.80 & 0.054 & 0.630 & 0.678 && 25.02 & 0.169 & 0.107 & 0.678\\
        & LayoutFlow    & 1.51 & 0.150 & 0.474 & 0.741 && 1.10 & 0.054 & 0.127 & 0.746\\
        \rowcolor{myhighlight} & VL-DE (ours)  & 2.92 & 0.154 & 0.539 & 0.532 && \rev{3.92} & \rev{0.089} & \rev{0.024} & \rev{0.435}\\
        \midrule
        \multirow{6}{*}{\taskrefinement} & RUITE & 7.93 & 0.177 & 0.492 & 0.658 && 7.89 & 0.073 & 0.038 & 0.637\\
        & LayoutFormer++   & 3.67 & 0.141 & 0.503 & 0.656 && 2.94  & 0.042 & 0.013 & 0.642\\
        \cmidrule{2-11}
        & LayoutDM$^*$      & 2.91 & 0.143 & 0.575 & 0.437 && 48.61 & 0.286 & 0.173 & 0.372\\
        & LayoutDiffusion   & 0.55 & 0.102 & 0.469 & 0.719 && 2.05  & 0.035 & 0.008 & 0.660\\
        & LayoutFlow        & 0.77 & 0.152 & 0.455 & 0.700 && 0.18  & 0.020 & 0.005 & 0.723\\
        \rowcolor{myhighlight} & VL-DE (ours)      & 0.84 & 0.136 & 0.454 & 0.688 && \rev{0.31} & \rev{0.046} & \rev{0.006} & \rev{0.721}\\
        \midrule
        & \textbf{Validation data} & 2.10 & 0.093 & 0.466 & 0.658 && 8.10 & 0.022 & 0.003 & 0.434\\
        \bottomrule
    \end{tabular}
    }
\end{table}

\subsection{Insertion Mechanism and Schedule} \label{app:rico-insertion}
All variants share the setup of Table~\ref{tab:hyperparameters-layout}, draw the category of a new element from a classifier, and insert following $\kappa(t)=\min(t/t_{\max},1)$. They differ in the inserted box:
\begin{itemize}[leftmargin=*, noitemsep, topsep=2pt]
    \item \textbf{Fully noised:} $x\sim\mathcal N(0,I)$, denoised on its own clock $u=(t-\tau)/(1-\tau)$.
    \item \textbf{Noisy data at the global time:} $x=t\,\hat x_1+(1-t)\,\varepsilon$, with $\hat x_1$ sampled from a learned head, denoised on the shared clock $t$.
    \item \textbf{Masked, then revealed:} a mask token, revealed later as in the previous bullet.
\end{itemize}
The first two predict one insertion rate per existing element, the third a single rate per layout. Since the layout is a set, the new element could be any missing one, so the target of the head is multimodal; we therefore compare a point estimate, a single Gaussian and a $16$-component Gaussian mixture.

Table~\ref{tab:rico-insertion} shows that per-element rates matter more than the form of the box, that $t_{\max}=0.5$ is best while $t_{\max}=1$ fails, and that the mixture head is best and converges in about half the epochs. The main results use the fully noised mechanism with $t_{\max}=0.5$.

\begin{table}[h]
    \centering
    \caption{\textbf{Insertion mechanism and schedule on RICO}, unconditional generation. Mechanism rows: $t_{\max}=0.5$, $5$ sampling seeds with the sampler of the main results. Schedule rows: masked-then-revealed mechanism, single runs. Head rows: the same model at fixed length, where all elements are present from the start and revealed from the head, single runs; ``epochs'' is the number of training epochs to reach a validation FID of $2.5$.}
    \label{tab:rico-insertion}
    \small
    \begin{tabular}{llrrrr}
        \toprule
        & Setting & \abbfid & \abbalignment & \abboverlap & epochs \\
        \midrule
        \multirow{3}{*}{Mechanism} & \cellcolor{myhighlight}Fully noised & \cellcolor{myhighlight}2.34 & \cellcolor{myhighlight}0.217 & \cellcolor{myhighlight}0.557 & \cellcolor{myhighlight}-- \\
        & Noisy data at the global time & 2.30 & 0.221 & 0.543 & -- \\
        & Masked, then revealed & 2.59 & 0.215 & 0.513 & -- \\
        \midrule
        \multirow{3}{*}{Schedule} & $t_{\max}=0.25$ & 2.55 & 0.240 & 0.497 & -- \\
        & $t_{\max}=0.5$  & 2.43 & 0.233 & 0.509 & -- \\
        & $t_{\max}=1$    & 11.94 & 0.269 & 0.645 & -- \\
        \midrule
        \multirow{3}{*}{Head} & Gaussian mixture ($16$ components) & 2.30 & 0.205 & 0.512 & 749 \\
        & Single Gaussian & 2.39 & 0.223 & 0.532 & 1524 \\
        & Point estimate  & 2.39 & 0.235 & 0.509 & 1299 \\
        \bottomrule
    \end{tabular}
\end{table}

\subsection{Other Ablations on RICO} \label{app:rico-ablations}
\emph{Clocks.} Training the category and box clocks independently, as Diffuse Everything does for any-to-any generation, costs almost two FID points; we train them jointly and obtain the conditional tasks from a task mixture instead ($2.58$ unconditional). Table~\ref{tab:rico-ablations} gives the numbers.

\begin{table}[h]
    \centering
    \caption{\textbf{Other ablations on RICO}, unconditional generation. $5$ sampling seeds.}
    \label{tab:rico-ablations}
    \small
    \begin{tabular}{llrrr}
        \toprule
        & Setting & \abbfid & \abbalignment & \abboverlap \\
        \midrule
        \multirow{2}{*}{Clocks} & Independent category and box clocks & 4.17 & 0.284 & 0.674 \\
        & Shared clock, task mixture & 2.58 & 0.169 & 0.497 \\
        \bottomrule
    \end{tabular}
\end{table}

\begin{revblock}
\paragraph{Hazard weight.} Applying Theorem~\ref{thm:loss-function} to the generator \eqref{eq:full-forward-process} would include the hazard rate $\lambda_t=\dot\gamma_t/(1-\gamma_t)$, which for $\gamma_t=\kappa(t)=\min(t/t_{\max},1)$ is $1/(t_{\max}-t)$ for $t<t_{\max}$ and diverges at the end of the insertion window; by Remark~\ref{rem:reweighting} any positive weight gives a valid loss. We compare three weights $w_t$ on the count and class terms of the insertion loss: $w_t=1$ (our loss), $w_t=\min(\lambda_t,50)$ and the same weight divided by its mean over $t\sim\mathcal U[0,1]$, $\ln 25+1\approx4.22$, so that it has the same average scale as the unweighted loss. All other settings are those of the main results with a single training seed shared by the three runs and $2{,}000$ epochs; we report the last checkpoint.

Table~\ref{tab:rico-hazard} shows that including the hazard weight worsens the FID and overlap. Normalizing it fixes this: at the end of training it matches the unweighted loss ($2.31$ against $2.29$ FID), which suggests that the overall scale of the weight hurts the performance rather than how it distributes the loss over time.

\begin{table}[h]
    \centering
    \caption{\textbf{Hazard weight on the insertion loss on RICO}, unconditional generation. Length TV is the total variation distance between the generated and test distributions of the number of elements. The unweighted row is a retrained copy of the model of the main results.}
    \label{tab:rico-hazard}
    \small
    \begin{tabular}{lrrrr}
        \toprule
        Insertion weight $w_t$ & \abbfid & \abbalignment & \abboverlap & Length TV \\
        \midrule
        \rowcolor{myhighlight} $1$ (ours) & $\mathbf{2.29}\pm0.13$ & $0.199$ & $0.560$ & $0.059$ \\
        $\min(\lambda_t,50)$ & $3.14\pm0.18$ & $0.305$ & $0.640$ & $0.056$ \\
        $\min(\lambda_t,50)/4.22$ & $2.31\pm0.12$ & $0.257$ & $0.536$ & $0.068$ \\
        \bottomrule
    \end{tabular}
\end{table}

\end{revblock}

\paragraph{Variable-length baselines.} All baselines share our backbone size, data and evaluation, and are reported with the best-by-validation checkpoint over $5$ sampling seeds. Edit Flows~\citep{havasi2025edit} and FlexMDM~\citep{kim2025any} run on the $32$-bin tokenization of \citet{chai2023layoutdm}, with $1{,}000$ steps; for FlexMDM we use the exact random unmasking order, and insert the $5$ tokens required per-box at once. OneFlow~\citep{nguyen2025oneflow} runs with $50$ steps. Branching Flows~\citep{billera2025branching} runs with $1{,}000$ steps and predicts the velocity toward its anchor, which was needed for the boxes to sharpen. \rev{To select the number of steps for each method we evaluated each baseline on a grid of $\{50,100,1000\}$ steps and reported the best result.}

\begin{figure}[h]
\centering
\ifdefined\arxivversion\def\qualwidth{0.85\linewidth}\else\def\qualwidth{\linewidth}\fi
\begin{subfigure}[t]{\linewidth}
    \centering
    \includegraphics[width=\qualwidth]{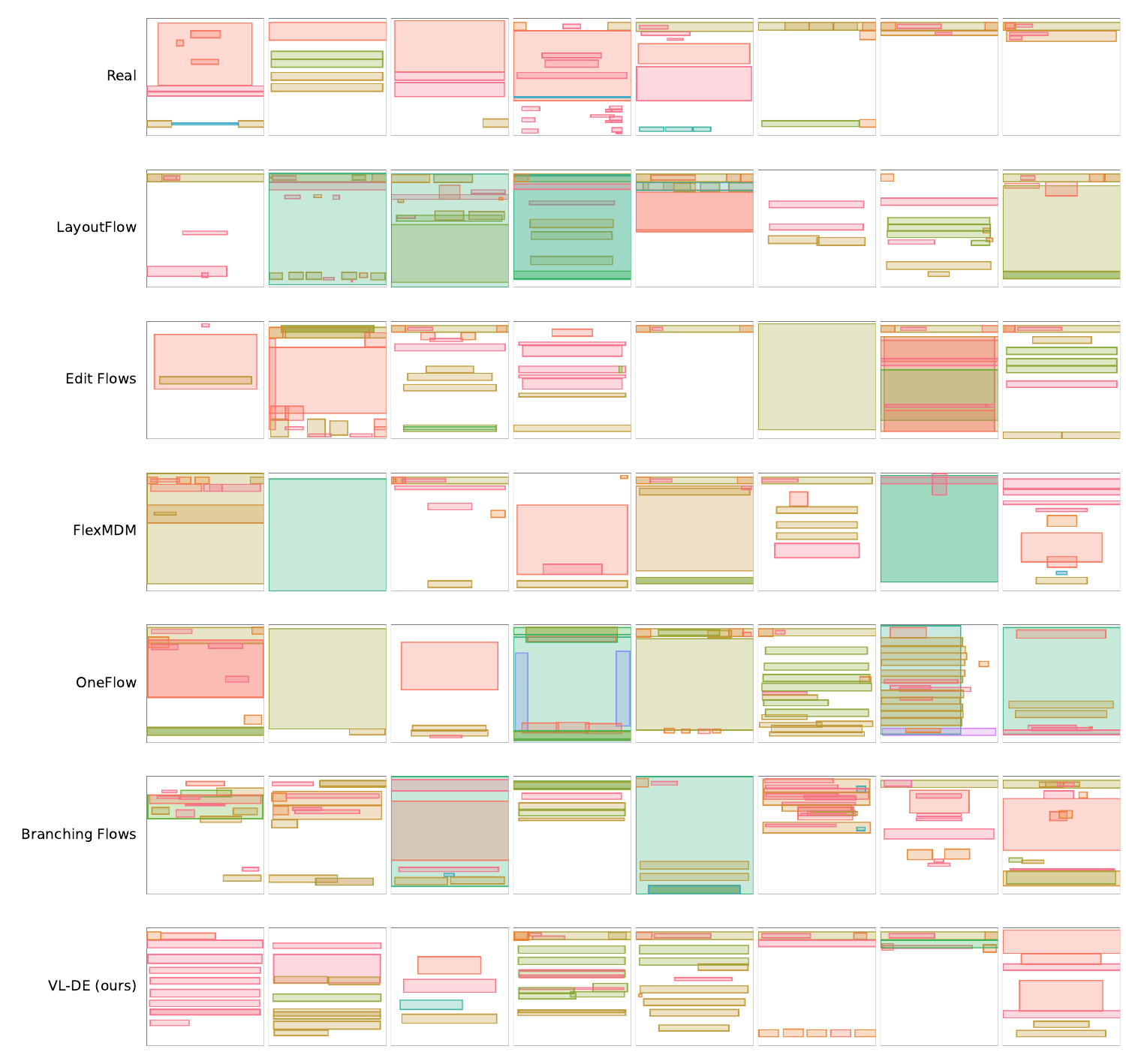}
    \caption{RICO ($25$ categories).}
\end{subfigure}\\[0.6em]
\begin{subfigure}[t]{\linewidth}
    \centering
    \includegraphics[width=\qualwidth]{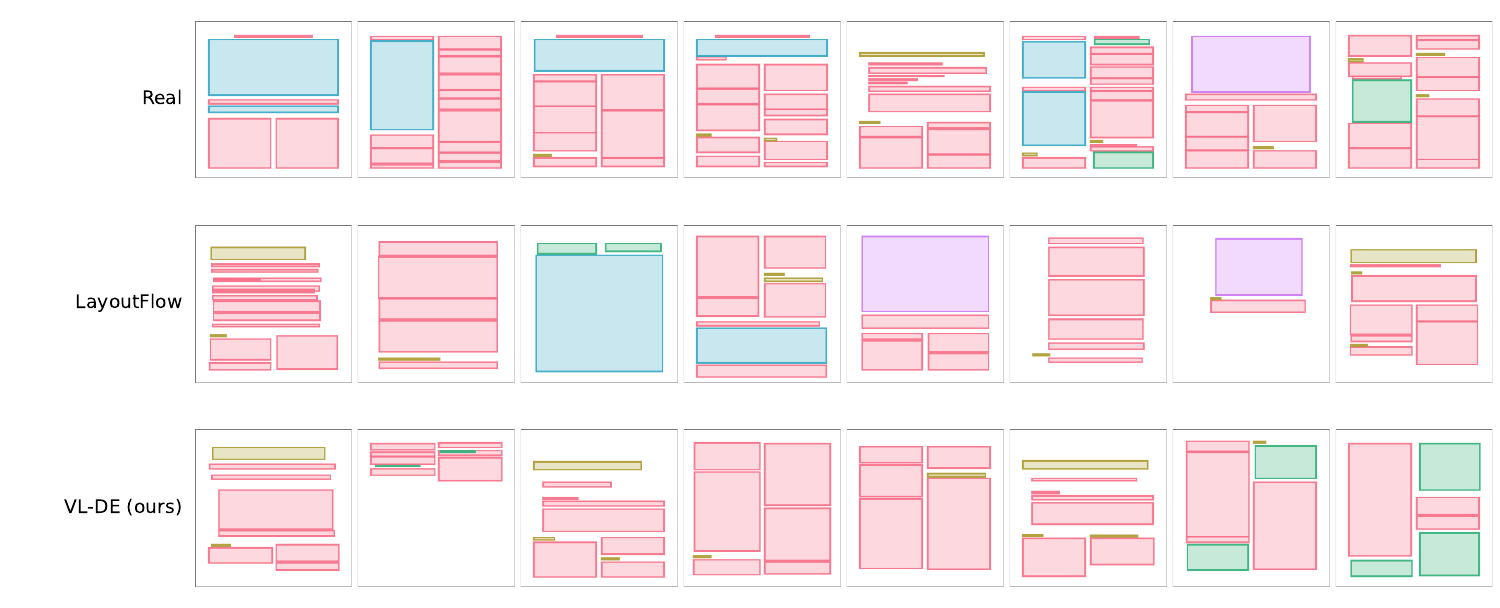}
    \caption{PubLayNet ($5$ categories: text, title, list, table, figure).}
\end{subfigure}
\caption{\textbf{Unconditional layout generation.} Rows are methods, columns eight layouts of a seeded batch; colors are categories. LayoutFlow is given the number of elements, the other methods generate it.}
\label{fig:layout-qualitative}
\end{figure}

\paragraph{Qualitative samples.} Figure~\ref{fig:layout-qualitative} shows samples of every method. LayoutFlow receives the number of elements; the other methods generate it.

\subsection{Content-Aware Layout Generation on CGL} \label{app:cgl-details}
\begin{figure}[h]
\centering
\includegraphics[width=\linewidth]{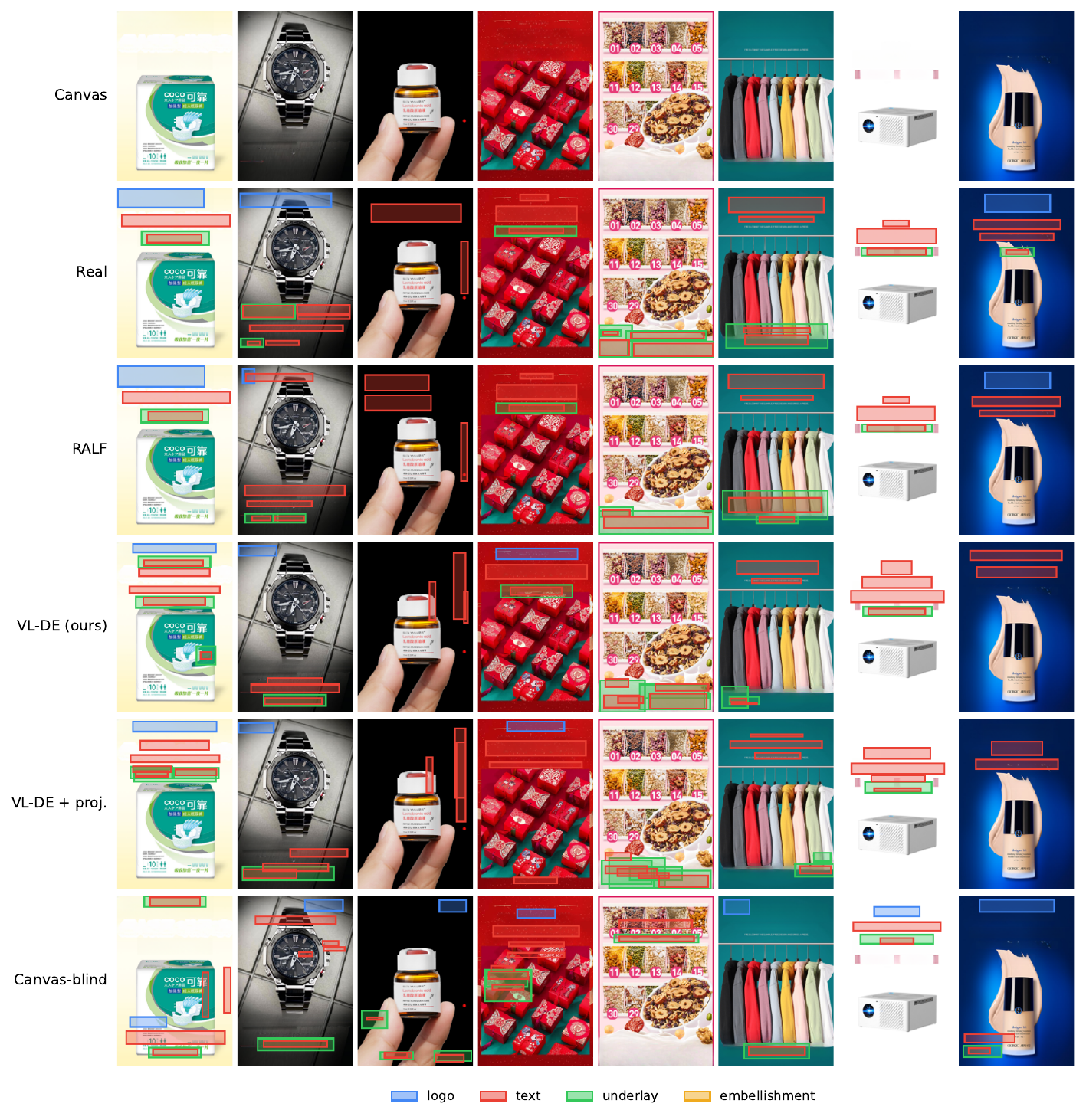}
\caption{\textbf{Content-aware layouts on CGL, unselected test canvases.} Boxes are drawn over the inpainted canvas; underlays are drawn beneath the other elements.}
\label{fig:cgl-qualitative}
\end{figure}

We use RALF's release of CGL~\citep{horita2024ralf} (splits of $48{,}544$ / $6{,}002$ / $6{,}002$ posters with at most $10$ elements, inpainted canvases and saliency maps) and run its evaluation script unmodified on our samples. The model is the RICO backbone of Table~\ref{tab:hyperparameters-layout} with the learned-initialization insertion, which transfers better than inserting from the prior on this dataset. Each canvas is encoded once by a frozen DINOv2-small~\citep{oquab2023dinov2} into an $8\times8$ grid of patch tokens with the mean saliency of each cell appended; elements read it through a gated cross-attention layer, and a pairwise geometric attention bias as in LayoutGD~\citep{layoutgd2026} is added between elements. To avoid memorizing the training canvases, we hide half of the canvas tokens and, for $10\%$ of the samples, the whole canvas, and flip canvas and layout horizontally with probability $0.5$. Training runs for $1{,}000$ epochs with the optimizer of Table~\ref{tab:hyperparameters-layout}, and the checkpoint is selected by validation FID. At the last sampling step, each underlay is enlarged to strictly contain the element that is already mostly inside it; this containment projection raises strict underlay effectiveness from $0.78$ to $0.95$ without changing FID. Table~\ref{tab:cgl-results-full} reports all of RALF's metrics and Figure~\ref{fig:cgl-qualitative} unselected samples.

\begin{table}[h]
    \centering
    \caption{\textbf{Content-aware layout generation on CGL} (test split, RALF's evaluator), grouped by how the number of elements is modeled. RALF rows are computed from their released outputs; $\dagger$ marks numbers taken from the papers, which report no FID. ``Proj.'' applies a single containment projection at the last sampling step (Appendix~\ref{app:cgl-details}). Table~\ref{tab:cgl-results} in the main text reports the four-column summary.}
    \label{tab:cgl-results-full}
    \small
    \resizebox{\linewidth}{!}{
    \begin{tabular}{lrrrrrr}
        \toprule
        Model & \abbfid & Occ$\downarrow$ & Rea$\downarrow$ & Und$_l\uparrow$ & Und$_s\uparrow$ & Ove$\downarrow$ \\
        \midrule
        \multicolumn{7}{l}{\textit{Histogram of $N$ (canvas ignored)}}\\
        VL-DE, canvas-blind                             & 1.63 & 0.318 & 0.041 & 0.960 & 0.877 & 0.006  \\
        \midrule
        \multicolumn{7}{l}{\textit{Fixed number of slots, padded with an empty class}}\\
        VL-DE backbone, padding                         & 4.02 & 0.144 & 0.022 & --    & 0.774 & 0.013  \\
        \midrule
        \multicolumn{7}{l}{\textit{Autoregressive with an end token}}\\
        RALF without retrieval                          & 2.89 & 0.125 & 0.019 & 0.967 & 0.917 & 0.011  \\
        \midrule
        \multicolumn{7}{l}{\textit{Insertion}}\\
        \rowcolor{myhighlight} VL-DE (ours)             & 1.53 & 0.148 & 0.023 & 0.933 & 0.784 & 0.0097 \\
        VL-DE + proj.                                   & 1.54 & 0.149 & 0.022 & 0.959 & 0.950 & 0.0097 \\
        \midrule
        \multicolumn{7}{l}{\textit{Reference: retrieval-augmented and published diffusion models}}\\
        RALF~\citep{horita2024ralf} (16 retrieved layouts) & 1.32 & 0.126 & 0.018 & 0.992 & 0.978 & 0.0042 \\
        LayoutDiT$^\dagger$~\citep{li2026layoutdit} & --   & 0.124 & 0.016 & 0.995 & 0.988 & 0.0016 \\
        LayoutGD$^\dagger$~\citep{layoutgd2026}        & --   & 0.120 & 0.014 & 0.999 & 0.994 & 0.0004 \\
        \midrule
        \textbf{Test data}                              & 0.80 & 0.125 & 0.017 & 0.995 & 0.988 & 0.0003 \\
        \bottomrule
    \end{tabular}
    }
\end{table}

\subsection{Demonstration: Independent Lengths on Crello} \label{app:crello-details}
\begin{figure}[h]
    \centering
    \includegraphics[width=0.85\linewidth]{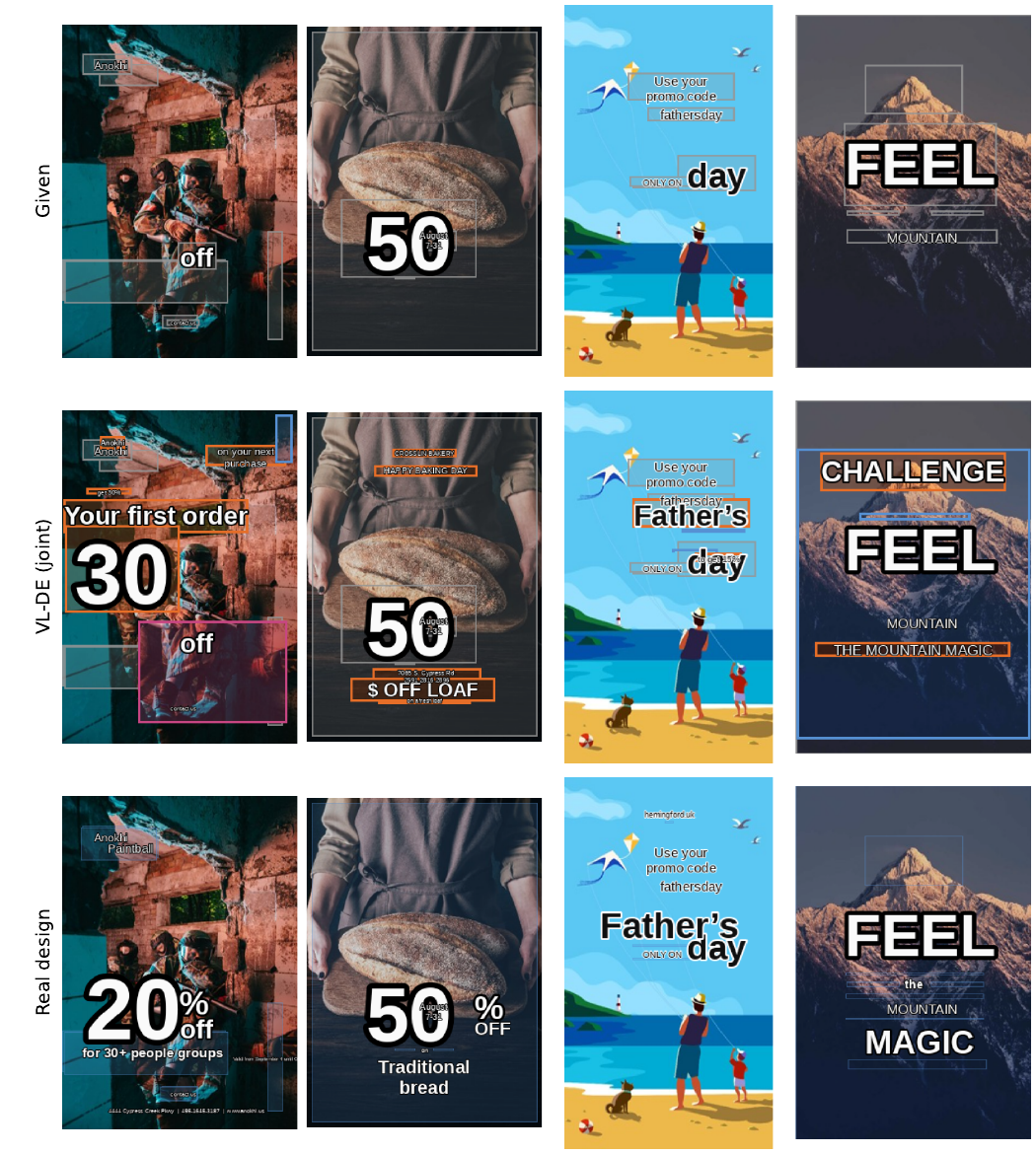}
    \caption{\textbf{Partial completion on Crello}, four test designs. \textbf{Top:} the given half of the design, in gray with its strings. \textbf{Middle:} the joint sampler of VL-DE adds elements (colored boxes) and writes their strings in the same run. \textbf{Bottom:} the real design. Designs were selected from test designs without a near-duplicate in the training set.}
    \label{fig:crello-samples}
\end{figure}

The per-element state may itself be variable-length. This section demonstrates that the construction accommodates two independently varying lengths in one sampling run; it is not a benchmark result. On Crello~\citep{yamaguchi2021canvasvae}, a design is a variable number of elements, and every text element carries a string of its own length. We keep the layout model of Section~\ref{sec:cgl-experiments}, which generates the number of elements by insertion, and give the strings their own generator, a fine-tuned autoregressive language model; which we model as a Markov process on the text coordinate, so it fits the construction unchanged, but it is not trained with our insertion objective. In partial completion, half of a design's elements are given with their strings, and the model must decide how many elements to add and write the strings of the new text elements. In the joint sampler the strings are written once during insertion, at $t=0.35$, and condition the remaining steps, so the element count and the string lengths are generated in one run. Compared with generating the layout first and the strings afterwards, this reduces the error in the number of added text elements from $1.26$ to $1.06$ (paired bootstrap interval $[0.12, 0.26]$ over $1{,}691$ test designs) and the layout FID from $2.48$ to $2.34$ (Table~\ref{tab:crello-partial}, samples in Figure~\ref{fig:crello-samples}). We also compare to a single autoregressive sequence over all elements and strings . We find that this approach models the lengths better, but at a large cost in layout quality. Extending this experiment to jointly generate the layout and the text that goes into them at a larger scale are left for future work, and hope this serves as a proof of concept. 

\begin{table}[!htb]
    \centering
    \caption{\textbf{Partial completion on Crello} ($1{,}691$ test designs; real designs add $4.4$ elements). Errors are mean absolute differences in the number of added elements and added text elements. $\dagger$: significantly better than the pipeline (paired bootstrap over designs).}
    \label{tab:crello-partial}
    \small
    \begin{tabular}{lrrr}
        \toprule
        Model & added-elem.\ err.$\downarrow$ & added-text err.$\downarrow$ & \abbfid \\
        \midrule
        Layout, then strings (pipeline)             & 2.09 & 1.26 & 2.48 \\
        \rowcolor{myhighlight} Joint (VL-DE)        & 1.96$^\dagger$ & 1.06$^\dagger$ & \textbf{2.34} \\
        Flat autoregressive sequence                & \textbf{1.81} & \textbf{0.89} & 8.54 \\
        \bottomrule
    \end{tabular}
\end{table}

\paragraph{Protocol.} We use the Crello dataset~\citep{yamaguchi2021canvasvae} ($18{,}414$ / $1{,}754$ / $1{,}869$ templates with at most $20$ elements, five element types). The first element of a template is the full-canvas background, which conditions the layout; the other elements are generated. The layout model is the CGL model of Appendix~\ref{app:cgl-details} with the font size of text elements as a fifth box coordinate and a per-element feature of the string (a CLIP text embedding and its character, word and line counts), trained with conditioning dropout and with batches in which a random subset of elements is given. The strings are written by Qwen2.5-7B-Instruct with a LoRA adapter, prompted with the template's title and keywords and the layout serialized as text, which writes all strings of a design in one pass and ends each at its own length. Layout FID uses a layout network trained on Crello with RALF's recipe.

For every test design with at least two text elements ($1{,}691$ designs), a random half of the elements is given with type, box, font size and string, with at least one text element given and one held out. The pipeline completes the layout without seeing the given strings and then writes the new strings. The joint sampler conditions on the given strings from the start, calls the language model once at $t=0.35$ on the current clean-layout estimate, and a last time for elements inserted afterwards. Rewriting the strings repeatedly during sampling is unstable, since each rewrite lengthens them and the layout grows to fit. The flat baseline is the same language model and adapter size trained to emit the whole design as one sequence, without the background image. Confidence intervals are paired bootstraps over designs with $5{,}000$ resamples.

\clearpage
\begin{revblock}
\subsection{Molecule Generation on QM9} \label{app:qm9-details}
\paragraph{Setup.} We train on QM9~\citep{ramakrishnan2014quantum}, we center every molecule so that its centroid is at the origin and apply a random rotation at each training step. The backbone is the EGNN of EDM~\citep{satorras2021n}, which predicts the clean position of every atom equivariantly; with a head on its last hidden state to predict the insertion rate and $q^\theta$ for every gap, as in \citet{campbell2023trans}. Additionally every atom receives its own clock $u(t;\tau)$ as input. We follow the evaluation protocol of \citet{billera2025branching} and report our hyperparameters in Table~\ref{tab:qm9-hyper}. 

\textbf{Samplers.} Figure~\ref{fig:qm9-samplers} reports the average $1-KS$ for our model under different step counts and network evaluations for our samplers. For each point we sample $2$K molecules, and find that when using less steps both of our samplers outperform the Euler sampler. On the other hand, when matching the network evaluations, we find that the split sampler lags behind in the beggining, with all samplers performing similarly under large NFEs.

\begin{table}[!ht]
    \centering
    \begin{revblock}
    \caption{Hyperparameters on QM9.}
    \label{tab:qm9-hyper}
    \setlength{\tabcolsep}{3pt}
    \begin{tabular}{ll}
        \toprule
        Backbone & EGNN, $9$ layers, width $256$ \\
        Head & Transformer, $4$ layers \\
        Parameters & 9.0M \\
        Insertion window & $t\in[0,0.5]$ \\
        Optimizer & AdamW, grad.\ clip $1$ \\
        Learning rate & $10^{-4}$, $500$ warmup steps \\
        Schedule & cosine to $10^{-6}$ \\
        Training steps & $1.35$M \\
        Batch size & $128$ \\
        EMA & $0.999$ \\
        Sampling & $1000$ Euler steps, EMA weights \\
        \bottomrule
    \end{tabular}
    \end{revblock}
\end{table}

\begin{figure}[!t]
    \centering
    \begin{revblock}
    \includegraphics[width=\linewidth]{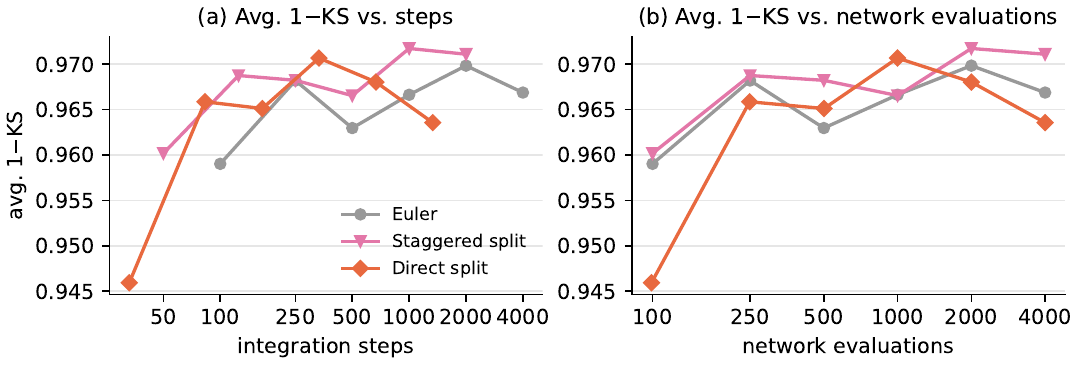}
    \caption{\textbf{Operator-splitting samplers on QM9} (average $1-$KS, $2$K molecules per point). (a) Under fixed step size our samplers outperform Euler (b) Per network evaluation, all samplers obtain similar performance, with the best results coming from our samplers demonstrating improved accuracy.}
    \label{fig:qm9-samplers}
    \end{revblock}
    \vspace{-\baselineskip}
\end{figure}

\end{revblock}

\begin{revblock}
\subsection{Arithmetic Equations} \label{app:arith-details}
In the other experiments every modality has the same length. Here each element is an arithmetic equation, which consists of a sequence of symbols and a sequence of numbers whose lengths differ and are linked only through the grammar (Figure~\ref{fig:arith-overview}a). VL-DE gives each modality its own insertion process (Section~\ref{sec:practical-process}).

\textbf{Data.} Each equation states that two random expressions have the same value; Figure~\ref{fig:arith-construction} illustrates how the data is generated. We start with a target value sampled uniformly from $[-10,10]$. We then randomly split each side of the equation with an operation ($+$, $-$ or $\times$): one part is a random expression, while the other is chosen to preserve the equality and is split in the same way until a single number remains. Parentheses only enclose sums that are multiplied, so none of them is redundant. Finally, we write the equation and separate it into the two sequences.

\begin{figure}[h]
    \centering
    \begin{revblock}
    \includegraphics[width=\linewidth]{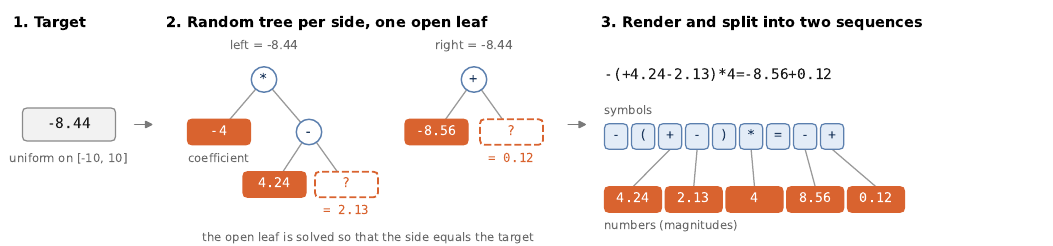}
    \caption{\textbf{Construction of an arithmetic equation.} A target value is drawn (1), each side is a random expression tree with one open leaf, which is solved so that the side equals the target (2), and the rendered equation is split into its symbols and the magnitudes of its numbers (3).}
    \label{fig:arith-construction}
    \end{revblock}
\end{figure}

\textbf{Dataset.} The symbols (the sign of every term, $\times$, parentheses and $=$) form one sequence, and the magnitudes of the numbers, in reading order, form the other. Equations have $2$ to $12$ numbers and $3$ to $25$ symbols, and $63\%$ of them contain parentheses. We use $20$K test equations and $2$M training equations.

\textbf{Model and training.} Both modalities share the global time and are inserted on $t\in[0,0.5]$. The MMDiT has one stream per modality joined by joint attention, receives the clock of every number, and predicts the clean numbers, the number of missing elements in every gap of each modality and the insertion distribution of the symbols. We use an MMDiT~\citep{esser2024scaling}. Hyperparameters are in Table~\ref{tab:arith-hyper}.

\textbf{Evaluation.} We sample $1000$ equations with $800$ Euler steps and parse the symbols with the grammar. An equation is valid if it can be parsed, and it is balanced if its left-hand and right-hand sides differ by less than $0.5$.

\textbf{Ablation: a single interleaved sequence.} To the best of our knowledge, no existing method covers the setup of multiple variable lengths. However, to obtain a better comparison, we implement a version of the dataset in which the data elements are presented as a single sequence, similar to \citet{nguyen2025oneflow}. We add a token $N$ that represents a number; when this token is selected, we insert Gaussian noise and denoise it on its own clock, as in VL-DE. 

\begin{table}[h]
    \centering
    \begin{revblock}
    \begin{minipage}[t]{0.6\linewidth}
    \centering
    \captionof{table}{Hyperparameters on the arithmetic equations.}
    \label{tab:arith-hyper}
    \scriptsize
    \setlength{\tabcolsep}{3pt}
    \begin{tabular}{lcc}
        \toprule
        & Ablation & Main model \\
        \midrule
        Training equations & $2$M & $2$M \\
        Depth (streams, joint) & $4$ & $8$ \\
        Width & $256$ & $256$ \\
        Parameters & 24.4M & 47.5M \\
        Training steps & $100$K & $900$K \\
        Optimizer & \multicolumn{2}{c}{AdamW, lr $10^{-4}$, $500$ warmup, clip $1$} \\
        Schedule & cosine to $10^{-6}$ & same, extended twice \\
        Batch size & $128$ & $128$ \\
        EMA & $0.999$ & $0.9999$ \\
        Sampling & \multicolumn{2}{c}{$800$ Euler steps, EMA weights} \\
        \bottomrule
    \end{tabular}
    \end{minipage}\hfill
    \begin{minipage}[t]{0.37\linewidth}
    \centering
    \captionof{table}{Valid samples and balanced valid samples (\%), $1000$ samples, $800$ Euler steps.}
    \label{tab:arith-results}
    \scriptsize
    \setlength{\tabcolsep}{10pt}
    \begin{tabular}{lrr}
        \toprule
        Model & Valid & Balanced \\
        \midrule
        \multicolumn{3}{l}{\textit{$100$K steps, depth $4$}}\\
        Interleaved & 92.9 & 24.9 \\
        \rowcolor{myhighlight} VL-DE & 91.9 & 20.5 \\
        \midrule
        \multicolumn{3}{l}{\textit{$900$K steps, depth $8$}}\\
        \rowcolor{myhighlight} VL-DE & 94.3 & 49.3 \\
        \bottomrule
    \end{tabular}
    \end{minipage}
    \end{revblock}
\end{table}
\textbf{Results.} Table~\ref{tab:arith-results} compares the two models at equal budget. We find that the interleaved model produces more valid and balanced equations. This is not surprising, since the interleaved sequence approach has more information, and it demonstrates that our approach is effective at learning such distributions. 

\end{revblock}

\end{document}